\documentclass[acmsmall]{acmart}
\usepackage{algorithm}
\usepackage[noend]{algpseudocode}
\newtheorem{theorem}{Proposition}
\newtheorem{assumption}{Assumption}
\usepackage{setspace}
\usepackage{multirow}
\usepackage{subfig}
\usepackage{graphicx}
\newcommand{\ie}{\textit{i}.\textit{e}.}
\newcommand{\eg}{\textit{e}.\textit{g}.}

\AtBeginDocument{%
  \providecommand\BibTeX{{%
    \normalfont B\kern-0.5em{\scshape i\kern-0.25em b}\kern-0.8em\TeX}}}

\setcopyright{acmcopyright}
\acmJournal{TOMM}
\acmYear{2021} \acmVolume{1} \acmNumber{1} \acmArticle{1} \acmMonth{1} \acmPrice{15.00}\acmDOI{}

\begin{document}

\title{Towards Corruption-Agnostic Robust Domain Adaptation}

\author{Yifan Xu}
\affiliation{%
  \institution{NLPR, Institute of Automation, Chinese Academy of Sciences {\&} School of Artificial Intelligence, University of Chinese Academy of Sciences}
  \streetaddress{95 East Zhongguancun Rd}
  \city{Beijing}
  \country{China}
  \postcode{100190}
}
\email{yifan.xu@nlpr.ia.ac.cn}

\author{Kekai Sheng}
\affiliation{%
  \institution{Youtu Lab, Tencent Inc.}
  \city{Shanghai}
  \country{China}}
\email{saulsheng@tencent.com}

\author{Weiming Dong}
\affiliation{%
  \institution{ NLPR, Institute of Automation, Chinese Academy of Sciences {\&} CASIA-LLvision Joint Lab}
  \city{Beijing}
  \country{China}
}
\email{weiming.dong@ia.ac.cn}

\author{Baoyuan Wu}
\affiliation{%
 \institution{The Chinese University of Hong Kong, Shenzhen; Shenzhen Research Institute of Big Data}
 \city{ShenZhen}
 \country{China}}
\email{wubaoyuan1987@gmail.com}

\author{Changsheng Xu}
\affiliation{%
  \institution{NLPR, Institute of Automation, Chinese Academy of Sciences {\&} School of Artificial Intelligence, University of Chinese Academy of Sciences}
  \city{Beijing}
  \country{China}}
\email{csxu@nlpr.ia.ac.cn}

\author{Bao-Gang Hu}
\affiliation{%
  \institution{NLPR, Institute of Automation, Chinese Academy of Sciences}
  \city{Beijing}
  \country{China}}
\email{ hubg@nlpr.ia.ac.cn}

\renewcommand{\shortauthors}{Xu, et al.}

\begin{abstract}
% 1. DA 目前的发展（套话）
% 2. 更真实且通用的场景——Corruption Robust Domain Adaptation (CRDA): 定义
% 3. 目前方法直接组合在CRDA上效果不好
% 4. 我们的思考：CRDA的两个挑战：
%     (1) 无标签的目标域
%     (2) 域差异损失约束力较弱
% 5. 由此，我们提出了 xxxxx 的方法：
%     (1) 我们采用基于inverting domain discrepency的方式构造worst-corruption sample 来隐式地表示unseen corruption; 并且将TSCL中的随机数据增广替换成worst-corruption sample。
%     (2) 为了解决训练早期特征不稳定的问题，我们采用teacher-student结构
% 6. 实验上看，我们的方法优于现有解决方案
% 7. 有趣的一点是：我们理论证明了只需要考虑最严重的severity就可以handle所有level severity，提升了方法效果和效率。
Big progress has been achieved in domain adaptation in decades. Existing works are always based on an ideal assumption that testing target domains are i.i.d. with training target domains. However, due to unpredictable corruptions (\eg, noise and blur) in real data like web images, domain adaptation methods are increasingly required to be corruption robust on target domains.
In this paper, we investigate a new task, Corruption-agnostic Robust Domain Adaptation (CRDA): 
% models are required 
to be accurate on original data and robust against unavailable-for-training corruptions on target domains. This task is non-trivial due to large domain discrepancy and unsupervised target domains.
We observe that simple combinations of popular methods of domain adaptation and corruption robustness have sub-optimal CRDA results.
% The challenges are two-fold: 1) unpredictable corruptions with large domain discrepancy; 2) weak constraints for robustness on unlabeled target domain.
We propose a new approach based on two technical insights into CRDA: 1) an easy-to-plug module called Domain Discrepancy Generator (DDG) that generates samples that enlarge domain discrepancy to mimic unpredictable corruptions; 
2) a simple but effective teacher-student scheme with contrastive loss to enhance the constraints on target domains. 
Experiments verify that DDG keeps or even improves performance on original data and achieves better corruption robustness than baselines.
% Interestingly, \textcolor{red}{we empirically find that networks always learn order-invariant representations for different severity levels of corruptions.} Based on this finding, we theoretically show only considering the most severe corruptions is enough for our framework to gain robustness on all levels of severity.
\end{abstract}



%   \includegraphics[width=\textwidth]{latex/figs/CRDA_Fig1_revisede.pdf}
%   \caption{Instances of corruption robust domain adaptation (CRDA).
%     %Here labeled clean source domain (art) and unlabeled clean target domain (real world) samples are available while training.
%     Existing domain adaptation method (CDAN+TN \cite{wang2019TransNorm}) and corruption robust method (AugMix \cite{hendrycks2019augmix}), even their combination, suffer from unseen corruptions on target domain.}
% %   \Description{Enjoying the baseball game from the third-base
% %   seats. Ichiro Suzuki preparing to bat.}
%   \label{fig:illustration}

\maketitle

%%%%%%%%% BODY TEXT
\section{Introduction}
% In the study of visual generalization, researchers mainly focus on robustness on domain shift and common corruptions. Domain shift is mainly studied by domain adaptation (DA) \cite{pan2009survey,long2018conditional,wang2019transferable,Li20DCAN,liang2020shot} and domain generalization (DG) \cite{volpi2018generalizing,dg_mmld,huang2020self} based on the scenario where labeled source domain samples (and unlabeled target domain samples for DA) are available while training to perform classification on target domain. Contributions to corruption robustness focus on the performance of the corrupted testing data (\eg, noise and blur) \cite{hendrycks2019benchmarking,hendrycks2019augmix,sun19ttt,hendrycks2019using}. However, researchers of DA or DG mainly consider domain shift on clean data, while current corruption robustness works focus on corruptions without domain shift. In a word, these two studies with a common goal for generalization \textcolor{red}{deliberately} ignored each other. Thus, there is a question to which both of them must face up: 
% 
% \textit{How to achieve both domain invariance and corruption robustness in a more generalized way?}
% 
\begin{figure*}
    \begin{center}
    \centering
    \includegraphics[width=\textwidth]{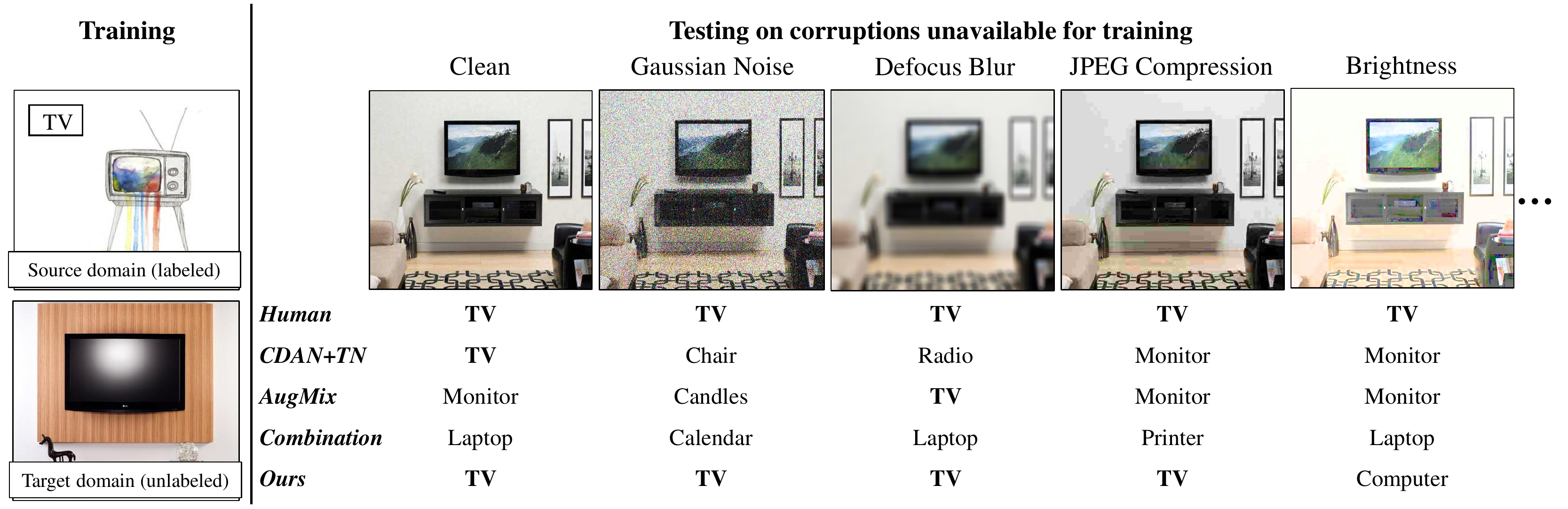}
    \captionof{figure}{Instances of corruption-agnostic robust domain adaptation (CRDA).
    %Here labeled clean source domain (art) and unlabeled clean target domain (real world) samples are available while training.
    To our surprise, existing domain adaptation method (CDAN+TN \cite{wang2019TransNorm}) and corruption robust method (AugMix \cite{hendrycks2019augmix}), even their combination, suffer from unseen corruptions on target domain.}
    \label{fig:illustration}
    \vspace{-3mm}
    \end{center}
\end{figure*}

Domain adaptation (DA)~\cite{pan2009survey} is a promising technique to transfer knowledge from well-labeled source domain to assist unlabeled target domain learning  with domain shift. Tremendous efforts on domain adaptation~\cite{Li20DCAN,liang2020shot,wang2019TransNorm,long2018conditional} and domain generalization (DG)~\cite{huang2020self,dg_mmld,volpi2018generalizing} indicate the significant progress on domain shift. Besides domain shift, given unpredictable corruptions (\eg, noise and blur) in real data, domain adaptation methods are increasingly required to be corruption robust on target domain (see Fig.~\ref{fig:illustration}). However, most DA or DG works only consider transferring source domain knowledge to some specific target datasets, while corruption robustness works~\cite{hendrycks2019benchmarking,hendrycks2019using,hendrycks2019augmix,sun19ttt} usually focus on corruptions without domain shift. Thus, there is a question worth considering: 
\textit{how to conduct robustness against unpredictable corruptions in cross domain scenarios?}

%%%%%%%%%%%%%%%%%%%%%%%%%%%%%%%%%%%%%%%%%%%%%%%%%%%%%%%%%%%%%%%%%%%
% \input{latex/tabs/differentDAsettings}

%Taking into consideration this question, we investigate a more realistic new task,
According to this question,  we propose a new task: \textbf{Corruption-agnostic Robust Domain Adaptation (CRDA)}, \ie, DA models are required to not only achieve high performance on original target domains, but also be robust against common corruptions that are \textbf{unavailable for training}.
%Popular methods of domain adaptation and corruption robustness, even their combinations, 
Popular DA methods, even combining with existing corruption robustness modules, will get sub-optimal results (see Fig.~\ref{fig:illustration}) because they cannot handle well two challenges of CRDA: 
%(1) Trade-off between domain invariance and corruption robustness. These two properties may conflict to each other due to \textcolor{red}{different optimization directions.}
(1) unpredictable corruptions with large domain discrepancy;
(2) weak constraints for robustness on unlabeled target domains. More specifically, previous augmentation robustness methods~\cite{hendrycks2019augmix,zhang2017mixup} for corruption-agnostic robustness are always conducted in supervised scenarios with strong classification loss, which may lose effectiveness with domain discrepancy loss on target domains. 
% reasonable utilization of domain information helps improving generalization to unseen corruptions. 
We first show that after taking into domain information, we can construct more generalized augmentation in cross-domain scenarios.
% Meanwhile, data augmentation may have no sense because it only encourages networks to memorize the specific corruptions seen during training and leaves models unable to generalize to new corruptions~\cite{geirhos2018generalisation,hendrycks2019augmix,vasiljevic2016examining}.
To our knowledge, our work is among the first attempt to unify robustness on domain shift and common corruptions.

To address the above challenges, we propose a novel mechanism towards corruption-agnostic robust domain adaptation called Domain Discrepancy Generator (DDG). Specifically, DDG generates augmentation samples that most enlarge domain discrepancy. Based on several assumptions (detailedly discussed in Section~\ref{sec:wc}),  these generated samples are proved to be able to represent unpredictable corruptions. Besides, to enhance the constraints on target domains and tackle with unstable features in the early training stage, we  propose a teacher-student warm-up scheme via contrastive loss. Specifically, a teacher model is first pre-trained to learn the original representations and then a student model further distills from the teacher model to learn robustness against samlpes generated by DDG via contrastive loss drawn from contrastive learning~\cite{chen2020simple}. Our code is available at \url{https://github.com/Mike9674/CRDA}.

Our work makes the following contributions:
\begin{itemize}
    \item We investigate a new scenario called corruption-agnostic robust domain adaptation (CRDA) to equip domain adaptation models with corruption robustness.
    
    \item We take use of information of domain discrepancy to propose a novel module Domain Discrepancy Generator (DDG) for corruption robustness that mimic unpredictable corruptions.
    
    \item Experiments demonstrate that our method not only significantly improves corruption robustness for DA models but also maintains or even improves classification results on original target domains.
    % \eg, the SOTA model DCAN achieves better performance with DDG.
\end{itemize}
%------------------------------------------------------------------------
\section{Related Work}
% \begin{table*}
%     \centering
%     \caption{Comparison of related research topics.}
%     \label{tab:DifferentDA}
%     \begin{tabular}{l|c|c|c}
%         \toprule
%         \textbf{Setting} & Domain Shift & Target Domain Available & Visual Corruption \\
%         \hline
%         Unsupervised Domain Adaptation & \checkmark & \checkmark &  \\
%         \hline
%         Domain Generalization & \checkmark &  &  \\
%         \hline
%         Corruption Robustness &  &  & \checkmark \\
%         \hline
%         Corruption Robust Domain Adaptation & \checkmark & \checkmark & \checkmark \\
%         \bottomrule
%     \end{tabular}
% \end{table*}

\begin{table*}
    \centering
    \caption{Comparison of related research topics.}
    \label{tab:DifferentDA}
    \resizebox{140mm}{!}{
    \begin{tabular}{l|c|c|c}
        \toprule
        \multicolumn{1}{c|}{\textbf{Setting}} & Domain Shift & Target Domain Available & Visual Corruption \\
        \hline
        Unsupervised Domain Adaptation~\cite{Li20DCAN,long2018conditional,pan2009survey} & \checkmark & \checkmark &  \\
        \hline
        Domain Generalization~\cite{huang2020self,dg_mmld,volpi2018generalizing} & \checkmark &  &  \\
        \hline
        Corruption Robustness~\cite{hendrycks2019benchmarking,hendrycks2019using,hendrycks2019augmix,shorten2019survey} &  &  & \checkmark \\
        \hline
        \textbf{Corruption-Agnostic Robust Domain Adaptation} & \checkmark & \checkmark & \checkmark \\
        \bottomrule
    \end{tabular}
    }
    \vspace{-3mm}
\end{table*}

\paragraph{Unsupervised Domain Adaptation.} 
% conclude mainstreams
% discuss and point out the differences of our paper
Tremendous DA methods have made progress in cross-domain applications like recognition~\cite{gopalan2011domain}, object detection~\cite{chen2018domain}, and semantic segmentation~\cite{tsai2018learning}. The core idea is to seek domain-invariant features among source and target domains~\cite{pan2009survey}. A mainstream methodology is distribution alignment, which is mainly based on Maximum Mean Discrepancy (MMD)~\cite{borgwardt2006integrating,chen2019joint,Li20DCAN,long2015DAN,pan2010TCA} or adversarial methods~\cite{ganin2016DANN,goodfellow2014GAN,long2018conditional,zhang2018collaborative,zhang2019domain}. Besides, some works further make improvement by pseudo-labeling~\cite{saito2017asymmetric}, co-training~\cite{zhang2018collaborative}, entropy regularization~\cite{shu2018dirt}, and evolutionary-based architecture design~\cite{sheng2021evolving}.
Recently, increasing researchers focus on more realistic scenarios: considering user privacy, \cite{li2020model,liang2020shot} investigate the scenario where only source domain models instead of data available while training. Label corruptions in source domain~\cite{han2020towards} is proposed to address the low quality labeling problem in DA. Besides, domain generalization~\cite{huang2020self,dg_mmld,volpi2018generalizing} aims to learn domain-invariant representations for unseen target domains.
Different from existing literature (see Table \ref{tab:DifferentDA}), we propose a new and realistic topic: corruption-agnostic robust domain adaptation, which investigates corruption robustness in domain adaptation.

% \vspace{-3mm}
\paragraph{Corruption Robustness.}
% conclude mainstreams
% \input{latex/tabs/differentDAsettings}
% discuss and point out the differences of our paper
Convolutional networks are proved fragile to simple corruptions by several studies~\cite{dodge2017study,hosseini2017google}. Assuming corruptions are known beforehand, Quality Resilient DNN~\cite{dodge2017quality} learns robustness against specific corruptions via a mixture of corruption-specific experts. 
% Zheng \textcolor{red}{et.al} aligns prediction of noisy images and clean ones to tackle underfitting on noise corruptions. 
Instead of knowing testing corruptions beforehand, we propose CRDA to learn general robustness against unseen corruptions. In recent years, increasing works begin to focus on robustness against unseen corruptions. \cite{vasiljevic2016examining} shows that fine-tuning on blurred images fails to generalize to unseen blurs. Several benchmarks~\cite{hendrycks2020many,hendrycks2019benchmarking,hendrycks2019nae,kang2019testing} are constructed to measure generalization to unseen corruptions. Self-supervised learning is found beneficial to corruption robustness~\cite{chen2020adversarial,hendrycks2019using}. CutMix~\cite{yun2019cutmix}, Mixup~\cite{zhang2017mixup}, Patch Gaussian~\cite{lopes2019improving}, Randaugment~\cite{cubuk2020randaugment} and AugMix~\cite{hendrycks2019augmix} are under the mainstream that aggregates several general transformations to implicitly represent unseen corruptions. However, current benchmarks and mainstream methods all take it for granted that training and testing data are from the same domain distribution, while CRDA also requires consideration of domain shift, as illustrated in Table \ref{tab:DifferentDA}. Furthermore, instead of aggregating transformations, we propose a new idea that utilize domain discrepancy information to mimic unseen corruptions.

% \paragraph{Adversarial Training}
% conclude mainstreams

% discuss and point out the differences of our paper
%------------------------------------------------------------------------
% \newpage

\section{Preliminaries}

%------------------------------------------------------------------------
\subsection{Problem definition for CRDA}
In this paper, we investigate CRDA in the scope  of unsupervised domain adaptation for classification. Given labeled source domain $\mathcal{D}_s = {\{\left(\mathcal{X}_{s},\mathcal{Y}_{s}\right)\}}$, unlabeled target domain $\mathcal{D}_t = {\{\mathcal{X}_{t}\}}$ ($\mathcal{Y}_{t}$ is unavailable while training), models are required to be robust on corrupted target domain samples $T(\mathcal{X}_{t})$ while maintaining the performance on original target domain, where $T$ denotes a corruption set \textbf{unavailable for training}. Note that $\mathcal{Y}_{s}=\mathcal{Y}_{t}=\{1,\ldots,S\}$ and $P\left(\mathcal{D}_s\right)\neq P\left(\mathcal{D}_t\right)$.

% \input{latex/tabs/frequent_notations}
%------------------------------------------------------------------------
\subsection{Definition of corruption robustness}
\label{sec:definition_robustness}
\begin{figure*}[t]
    \centering
    \includegraphics[width=0.7\textwidth]{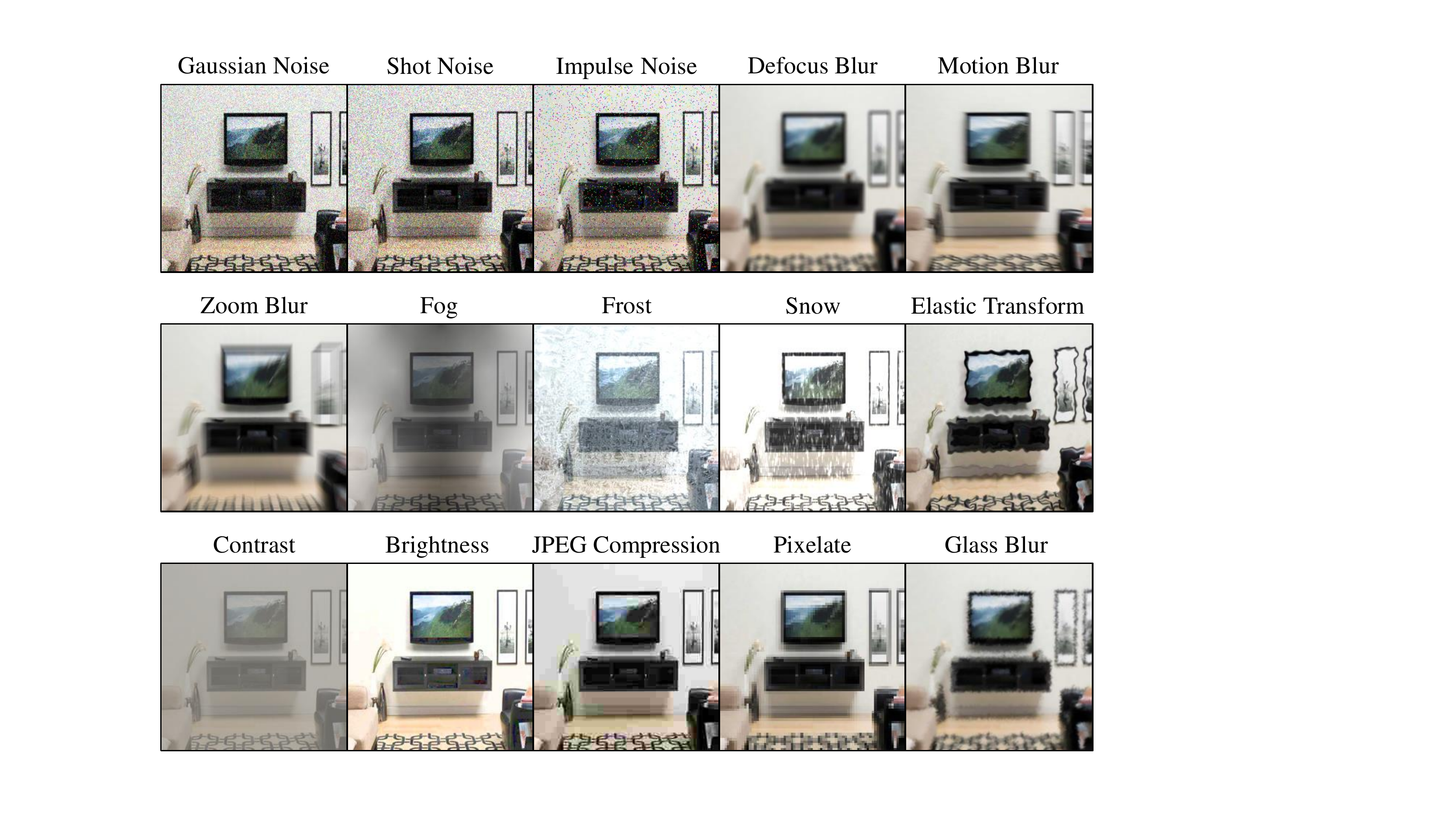}
    \caption{Examples of $15$ kinds of corruptions.}
    \label{fig:15corruption1}
\end{figure*}

The concept of corruption robustness is drawn from \cite{hendrycks2019benchmarking} with slight modification. Given an input sample set $\mathcal{X}$ with the corresponding label set $\mathcal{Y}$, a corruption set $T$ and a model $m=f \circ c$, where $f$ and $c$ respectively denote a feature extractor and a classifier, the corruption robustness is measured by:
\begin{equation}
\mathbb{E}_{t \sim T}\left[\mathbb{P}_{(x, y) \sim (\mathcal{X},\mathcal{Y})}(m(t(x))=y)\right],
% \forall (x,y) \in (X,Y), t \in T: m(t(x))=m(x)=y
\label{ideal_transforamtion_definition}
\end{equation}
% In practice, if \textcolor{red}{identity transformation} lies in corruption set $T$, Equation (\ref{ideal_transforamtion_definition}) is always achieved by:
% \begin{equation}
% \min _{\theta} E_{t \in T}\left[E_{(x,y) \in (X,Y)} \ell(t(x),y, \theta)\right]
% \end{equation}
% where $\ell$ denotes an \textcolor{red}{effective} loss function.
%
%In this paper, we derive Equation (\ref{ideal_transforamtion_definition}) by aligning features:
% In this paper, we derive the robustness defined in Equation (\ref{ideal_transforamtion_definition}) by aligning features, which is shown in Equation (\ref{ideal_transforamtion_definition}).
where we can derive the robustness by aligning features as:
\begin{equation}
\max \left\{\mathbb{E}_{t \sim T}\left[\mathbb{P}_{x \sim \mathcal{X}}(f(t(x))=f(x))\right]\right\}.
% \forall x \in X, t \in T: f(t(x))=f(x)
\label{practice_transforamtion_definition}
\end{equation}
Here $T$ denotes corruptions derived from \cite{hendrycks2019benchmarking}, which contains 15 kinds of corruptions for $T$ and 5 severities $t$ for each kind (see Fig.~\ref{fig:15corruption1}).
% The total  15 kinds of corruptions include: Gaussian Noise, Shot Noise, Impulse Noise, Defocus Blur, Motion Blur, Zoom Blur, Fog, Frost, Snow, Elastic Transform, Contrast, Brightness, JPEG Compression, Pixelate and Glass Blur.
% Some corruption examples are shown in Figure \ref{fig:illustration}, details of all kinds of corruptions and severities are shown in SM.
% Consider Spatial Transformation Later

%------------------------------------------------------------------------
% 1. Domain Discrepancy Generator

% 2. Overall learning framework
% 补充一个公式：l_ori包括分类损失和迁移损失。
% 3. Further explanation

% 4. Metrics
% 要突出我们自己发布了生成函数（类似Benchmark），并附图

\section{Methodology}
% In this paper, our attempts are made through two key problems of CRDA: 1) trade-off between domain invariance and corruption robustness; 2) generalization to unseen corruptions via domain information. we first propose a simple but effective teacher-student scheme (TSCL) for the first problem. As for the second problem, we further construct worst-corruption samples to mimic unseen corruptions via inversely enlarging domain discrepancy. Putting them together, we derive \textbf{TSCL-WC} as illustrated in Figure \ref{fig:AdvTL}. In addition, we theoretically show that only considering the most severe level of corruptions is  enough for our frameworks to handle all levels of severity.

%----------------------------------------------------------------------------------------------------------
\subsection{Domain Discrepancy Generator}
\label{sec:wc}

\begin{figure}
    \begin{center}
    \includegraphics[width=0.6\textwidth]{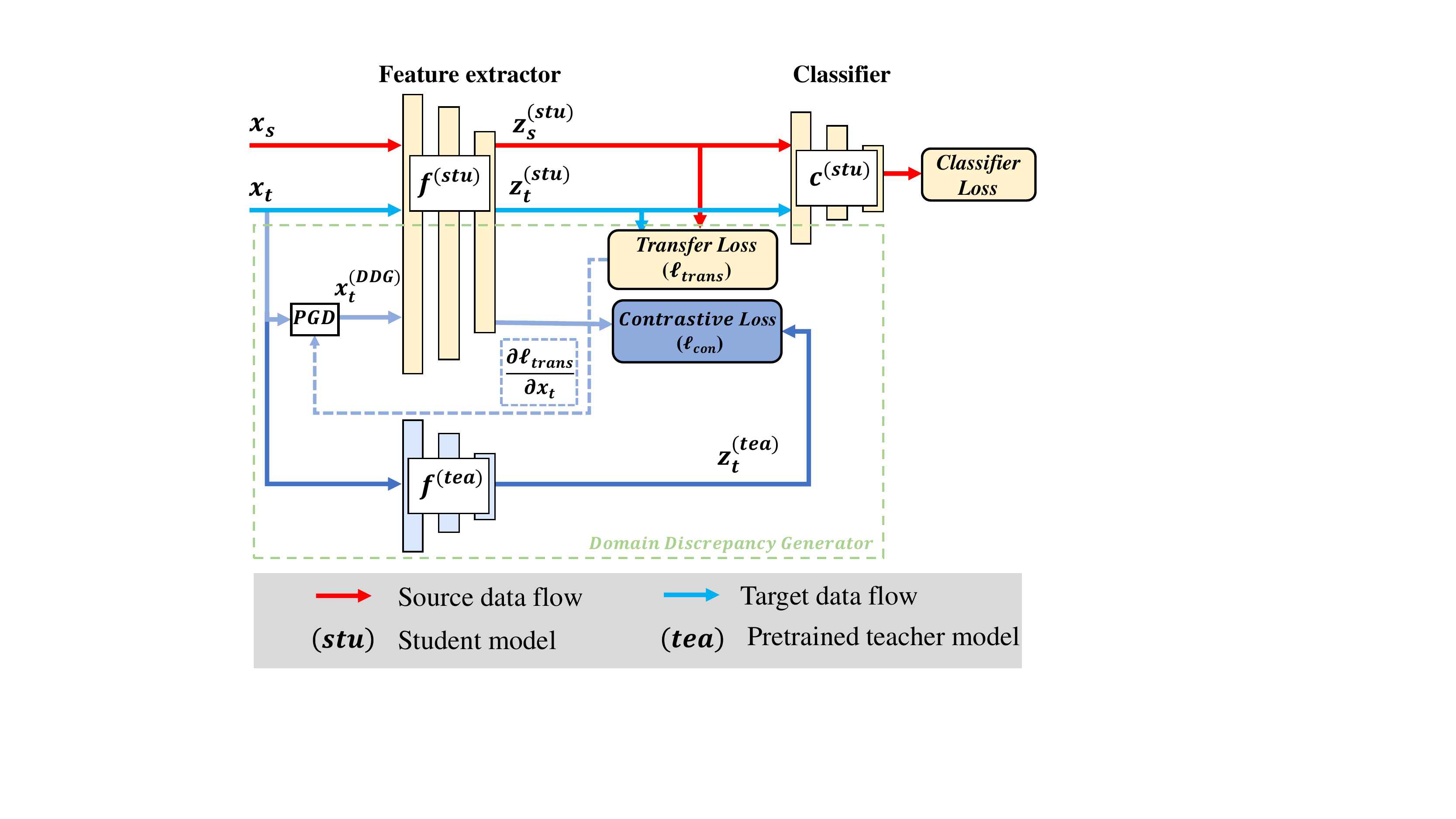}
    \end{center}
    \caption{Overall structure of DDG. The yellow part is the classic structure of DA, while the green part is DDG. The gradient of transfer loss is inversely added to the target domain inputs by Projected Gradient Descent (PGD)~\cite{madry2017PGD} in Algorithm \ref{algorithm-PGD} to generate augmentation samples $x_{t}^{(DDG)}$ which is aligned with the original domain-invariant feature extracted by a pretrained teacher model with contrastive loss.}
    \label{fig:AdvTL}
    \vspace{-3mm}
\end{figure}

In CRDA setting, the key challenge is that testing corruptions are unavailable during training. Besides, previous data augmentation only encourages networks to memorize the specific corruptions seen during training and leaves models unable to generalize to new corruptions~\cite{geirhos2018generalisation,vasiljevic2016examining}. In this section, we attempt to utilize domain information given by domain adaptation to solve this challenge.
Thus, Domain Discrepancy Generator (DDG) is proposed to generate samples that most enlarge domain discrepancy to mimic unpredictable corruptions. Fig.~\ref{fig:AdvTL} illustrated the overall mechanism of DDG. The following is reasonable derivation.

We first assume that corrupted target domain samples hold larger domain discrepancy to source domain than corresponding original target domain samples, which is empirically proved in the supplementary material. 
Thus, our core idea is to find the points near the original images in sample space that most increase domain discrepancy to represent the most severe corruptions. 
In domain adaptation, the domain discrepancy is always measured by transfer loss, which is commonly realized by Maximum Mean Discrepancy (MMD)~\cite{borgwardt2006integrating,Li20DCAN,long2015DAN,pan2010TCA} or adversarial methods \cite{ganin2016DANN,long2018conditional}. 

The followings are three basic assumptions for our method, where distance $d_{*}(t(x),x)$ denotes distance between the corrupted images and corresponding original images:
\begin{assumption}
\label{assum:assum1}
The corrupted versions are within the $\delta$-range neighborhood of the original image in sample space ($\boldsymbol{\delta}$ \textbf{neighborhood} for precise) and distance in sample space $d_{s}(t(x),x)$ becomes larger with increase of the severity level of corruption $t$.
\end{assumption}
\begin{assumption}
\label{assum:assum2}
Within the $\delta$ neighborhood, the severity level $t$ of a corruption  $T$ is positively correlated to distance in feature space  $d_{f}(f(t(x)) ,f(x))$.
\end{assumption}
\begin{assumption}
\label{assum:assum3}
Within the $\delta$ neighborhood, the severity level $t$ of a corruption $T$ is positively correlated to the value of transfer loss $\ell_{trans}(t(x))$.
\end{assumption}

The empirical proof can be seen in supplementary material. Assumption~\ref{assum:assum2}  is drawn from an interesting finding in Section~\ref{sec:empirical} that networks always learn order-invariant representations for different severity levels of corruptions.

Based on the above assumptions, we can construct the bridge between domain discrepancy and unpredictable corruptions. Given an input target domain data $x_{t}$, source domain $D_{s}$, transfer loss $\ell_{trans}$ and a corruption set $T$ ($t \in T$) whose corresponding corrupted data $t(x_{t})$ is within the $\delta$ neighborhood of $x_{t}$, DDG generates
\begin{equation}
    \begin{split}
    x_{t}^{(DDG)}&=\mathop{\arg\max}_{\left\|x_{t}^{(DDG)}-x_{t}\right\| \leq \delta} \ell_{trans}\left(f(x_{t}^{(DDG)}), f(D_{s})\right)
    \\&\Rightarrow \mathop{\arg\max}_{t \in T} \ell_{trans}\left(f(t\left(x_{t})\right), f(D_{s})\right)=t_{\max }(x_{t}),
    \end{split}
    \label{construction_x_adv}
\end{equation}
which means the point $x_{t}^{(DDG)}$ near the original target domain sample in sample space that most increases transfer loss (denoting domain discrepancy) can represent the most severe corruption. The proof of Equation (\ref{construction_x_adv}) can be seen in the supplementary material.
In practice, Domain Discrepancy Generator generates such samples via Project Gradient Descent (PGD)~\cite{madry2017PGD} in Algorithm \ref{algorithm-PGD}.
\begin{algorithm}[t]
    \caption{PGD: Projected Gradient Descent}
    \small{
    \textbf{Input:} 
    \begin{minipage}[t]{0.87\linewidth}
    Target and source doamin data $x_{t},x_{s}$; Feature  extractor $f$;Parameters: shift range $\delta$, update stride $\eta$, update step number $n$.
    \end{minipage}
    
    \textbf{Output:}
    \begin{minipage}[t]{0.86\linewidth}
    augmented target samples by DDG $x_{t}^{(DDG)}$
    \end{minipage}
    
    \begin{algorithmic}[1]
    
    \State $i=0$, $x_{t}^{(DDG)}=x_t$
    \Repeat :
    \State Transfer loss:
    
    $\ell=\ell_{trans}(f(x_{t}^{(DDG)}),f(x_{s}))$

    \State Update $x_{t}^{(DDG)}$ by the gradients of transfer loss:
    
    $x_{t}^{(DDG)} \leftarrow x_{t}^{(DDG)} + \eta \cdot \text{sign} \left(\frac{\partial {\ell}}{\partial x_{t}^{(DDG)}}\right)$
    
    \State Clamp $x_{t}^{(DDG)}$ within $\left\|x_{t}^{(DDG)}-x_{t}\right\| \leq \delta$
    \State $i \leftarrow i+1$
    \Until $i=n$
    \State \Return $x_{t}^{(DDG)}$
    
    \end{algorithmic}
    \label{algorithm-PGD}
    }
\end{algorithm}

It is worthy noting that the generated samples look similar to adversarial samples~\cite{madry2017PGD}. However, there exists few evidence that domain discrepancy could help corruption robustness in cross domain scenarios like domain adaptation ever before. We first show that it significantly improves corruption robustness on the basis of the bridge between corruptions and domain discrepancy constructed by Equation (\ref{construction_x_adv}). Meanwhile, DDG works in an unsupervised way while common adversarial training always needs ground-truth labels.

\subsection{Overall Learning Framework} 
The last question is how to learn robustness in unlabeled target domains. Since there is no strong constraints like classification loss in target domains, simply merging samples generated by DDG with the original data may lose the effectiveness. Thus, we utilize contrastive loss~\cite{chen2020simple} to enhance the constraints on target domains. The core idea is simple, \ie, minimizing the feature distance between corrupted samples with their original versions. Besides, to tackle with the unstable features in the early training stage, we further propose a warm-up scheme like teacher-student framework. The teacher model is first trained to extract the original features, while the student model then extracts corrupted features to minimize the distance to corresponding original features. Algorithm~\ref{algorithm-DDG} shows the overall process of DDG. The proposed contrastive loss is introduced as followed.

\begin{algorithm}[t]
    \caption{Overall process for Domain Discrepancy Generator}
    \label{algorithm-DDG}
    \small{
    \textbf{Input:} 
    \begin{minipage}[t]{0.87\linewidth}
    Data and source labels: $\mathcal{X}_{s}$, $\mathcal{X}_{t}$, $\mathcal{Y}_{s}$; Original DA model $f+c$; Parameters: $\delta=60/255$, $\eta > 2\delta$, $n=2$
    \end{minipage}
    
    \textbf{Output:} 
    \begin{minipage}[t]{0.86\linewidth}
    Robust student model $f^{(stu)}+c^{(stu)}$
    \end{minipage}
    
    \begin{algorithmic}[1]
    \State Train the original model as a teacher model $f^{(tea)}+c^{(tea)}$. Fix the parameters.
    \State 	Construct a student network $f^{(stu)}+c^{(stu)}$ with same structure of $f+c$. 
    \Repeat :
    \State Draw $x_{s}, x_{t}, y_s$ from $\mathcal{X}_{s}, \mathcal{X}_{t}, \mathcal{Y}_{s}$.
    \State 
    \begin{minipage}[t]{0.9\linewidth}
    Generate augmentation samples according to Equation (\ref{construction_x_adv}) and Algorithm \ref{algorithm-PGD}:
    
    $\qquad x_{t}^{(DDG)}=\text{PGD}\left(x_{t},x_{s},f^{(stu)},\delta, \eta, n\right)$
    \end{minipage}
    
    \State
    Update $f^{(stu)}+c^{(stu)}$ according to Equation (\ref{total_loss}).
    
    \Until Convergence
    \State \Return $f^{(stu)}+c^{(stu)}$
    
    \end{algorithmic}
    }
\end{algorithm}

Given a batch of samples ${\{x_i\}}_1^N$, we use feature extractor $f^{(tea)}$ and $f^{(stu)}$ introduced in Algorithm \ref{algorithm-DDG} to get $2N$ representation feature vectors:
\begin{equation}
    \begin{split}
        Z^{(tea)}={\left\{z_{i}^{(tea)}|z_{i}^{(tea)}=f^{(tea)}(x_i)\right\}}_1^N,\\
        Z^{(stu)}={\left\{z_{i}^{(stu)}|z_{i}^{(stu)}=f^{(stu)}(x_i)\right\}}_1^N,
    \end{split}
\end{equation}
% $Z^{(tea)}={\{z_{i}^{(tea)}|z_{i}^{(tea)}=f^{(tea)}(x_i)\}}_1^N$ and $Z^{(stu)}={\{z_{i}^{(stu)}|z_{i}^{(stu)}=f^{stu}(x_i)\}}_1^N$. 
where we define $Z=Z^{(tea)}\cup Z^{(stu)}={\{z_i\}}_1^{2N}$. The similarity loss of each two features can be calculated by:
\begin{equation}
    \ell_{sim}\left(z_{i}, z_{j}\right)=-\log \frac{\exp \left(sim\left(z_{i}, z_{j}\right) / \tau\right)}{\sum_{z_{k} \in Z, z_{k} \neq z_{i}} \exp \left(sim\left(z_{i}, z_{k}\right) / \tau\right)},
    \label{similarity}
\end{equation}
where $\tau=0.2$ is a temperature controller to control the similarity extent, and $sim\left(z_i,z_j\right)={z_i}^T z_j/z_i z_j$ is cosine distance between two vectors. 
% which is $sim\left(z_i,z_j\right)={z_i}^T z_j/z_i z_j$. 
Note that Equation (\ref{similarity}) is asymmetric between $z_i, z_j$. The final contrastive loss is defined by:
\begin{equation}
    \begin{split}
    \ell_{con}\left(Z^{(stu)}, Z^{(tea)}\right)
    = \frac{1}{2 N} \sum_{i=1}^{N} \left[\ell_{sim}\left(z_{i}^{(stu)}, z_{i}^{(tea)}\right)+ \ell_{sim}\left(z_{i}^{(tea)}, z_{i}^{(stu)}\right)\right],
    \end{split}
    \label{CL}
\end{equation}
where the goal is to minimize the distance between feature representations $z_{i}^{(stu)}$ and $z_{i}^{(tea)}$ of a same sample $x_i$. The other kinds of losses defined by the original DA models $\ell_{ori}$, usually containing source domain classification loss $\ell_{cls}$ and transfer loss $\ell_{trans}$in Equation (\ref{ori_loss}),  also need to be calculated:
\begin{equation}
\label{ori_loss}
    \ell_{ori}(x_{s},x_{t},y_{s})=\ell_{cls}(x_{s},y_{s})+\ell_{trans}(x_{s},x_{t}).
\end{equation}

In fact, contrastive loss does not make $Z^{(stu)}$ definitely same as $Z^{(tea)}$. And thanks to corruptions, the final feature representation $Z^{(stu)}$ has a further distillation on the basis of the teacher model, which may lead to not only improvement on robustness but also better performance on domain invariance.

By utilizing the contrastive loss in Equation (\ref{CL}), we can minimize feature distance between samples $x_{t}^{(DDG)}$ generated by DDG and original samples $x_{t}$ in the target domain iteratively, which leads to features of the most severely corrupted images come closer to original ones, as:
\begin{equation}
    \min d_{f}\left(f(x_{t}^{(DDG)}), f(x_{t})\right) \Rightarrow \min d_{f}\left(f(t_{max}(x_{t})), f(x_{t})\right).
    \label{illustration1}
\end{equation}
In practice, Equation (\ref{illustration1}) is realized by minimizing:
\begin{equation}
    \ell_{con}^{(DDG)}\left(x_{t}^{(DDG)},x_{t}\right)=\ell_{con}\left(f^{(stu)}(x_{t}^{(DDG)}),f^{(tea)}(x_{t})\right)
    \label{CL-AdvTL},
\end{equation}
Together with the loss defined by the original DA model $\ell_{ori}$ illustrated in Equation (\ref{ori_loss}) and Fig.~\ref{fig:AdvTL}, the final total loss is:
\begin{equation}
    \ell_{total}=\ell_{ori}(x_{s},x_{t},y_{s})+\lambda \ell_{con}^{(DDG)}(x_{t}^{(DDG)},x_{t}).
    \label{total_loss}
\end{equation}

Note that DDG only generated samples on target domains and the generated samples are only processed by contrastive loss.

%-----------------------------------------------------------------------------------------
\subsection{Further Explanation}
\label{sec:theory}
Intuitively, DDG should loop many times like $n=10$ in PGD to generates an augmented sample, which is time-consuming. In this section, we theoretically show that we only need to consider the edge point so that by setting $n=2$ is enough, which effectively reduces the time consumption. We begin with the following proposition.

\begin{theorem}
Only aligning the edge points $\{x_{t}^{(DDG)}|\|x_{t}^{(DDG)}-x_{t}\|=\delta\}$ around the $\delta$ neighborhood in Equation (\ref{construction_x_adv}) is enough to gain corruption robustness under the DDG framework.
\label{pro2}
\end{theorem}
 
Given input data $x$, a feature extractor $f$ and a corruption $T$ with continuous severity, we denote the distance in feature space $d_{f}(f(T(x)),f(x))$ as $d(T)$. Then Assumption~\ref{assum:assum2}  can be explained as  $T$ is positively correlated with $d(T)$. Due to the properties of monotonic function, the upper bound of $d(T)$ is $d(T_{max})$:
 \begin{equation}
    0 \leq d(T) \leq d(T_{max}) = d_{max},
    \label{max-dT}
\end{equation}
where $T_{max}$ denotes the most severe level of corruption $T$.
By limiting $d_{max}$ to $0$, we get:
\begin{equation}
    \lim_{d_{max} \rightarrow 0} {d(T)}= \lim_{d_{max} \rightarrow 0} {d_{f}(f(T(x)),f(x))}=0,
\label{limit}
\end{equation}
which means for all severity levels of corruption $T$, features of corrupted data $T(x)$ and clean data $x$ are the same when $d_{max} \rightarrow 0$. Thus, robustness against a specific corruption is achieved.

Now the last issue is how to limit $d_{max}$ to 0. Note that:
\begin{equation}
    d_{max}=d(T_{max})=d_{f}(f(T_{max}(x)),f(x)).
    \label{dmax}
\end{equation}
Thus, by aligning features of the most severely corrupted samples $T_{max}(x)$ with original features $x$, we get $d_{max}\rightarrow0$. 

Considering Assumption~\ref{assum:assum1}, the shift in sample space increases as the increase of the severity level of a corruption $T$. Thus, the most severe corruption $T_{max}(x)$ is always achieved on the edge points of $\delta$ cycle. Then we can use $\{x_{t}^{(DDG)}|\|x_{t}^{(DDG)}-x_{t}\|=\delta\}$ to implicitly represent $T_{max}(x)$ due to the conclusion in Equation (\ref{construction_x_adv}).  By aligning features of these edge points with the original features via minimizing the contrastive loss defined in Equation (\ref{CL-AdvTL}), corruption robustness is achieved under DDG as :
\begin{equation}
    \lim_{\ell_{con}^{(DDG)} \rightarrow 0} {d(T)}= \lim_{\ell_{con}^{(DDG)} \rightarrow 0} {d_{f}(f(T(x)),f(x))}=0.
    \label{limit_xadv}
\end{equation}
Proposition \ref{pro2} is derived.

Thus, by setting stride $\eta$ bigger than range $2\delta$, we can reach the edge point in one step. In practice, only step $n=2$ gains enough good performance.
% In practice, results shows that 
% \textcolor{red}{Based on Assumption 2, we conclude the result that } 
% In practice, we \textcolor{red}{are surprised to} find that only aligning the edge points $\{x_{t}^{(wc)}|\|x_{t}^{(wc)}-x_{t}\|=\delta\}$ in TSCL-WC gets best performance, which is realized by setting $\eta > 2\delta$ in \textcolor{red}{Algorithm}. Intuitively, the setting with $\eta \ll \delta$ seems like a more reasonable setting but gets poorer performance on large shift transformations like contrast. 

%-----------------------------------------------------------------------------------------
\subsection{Metrics for corruption robustness}
The commonly used standardized aggregate performance measure is the Corruption Error (CE) \cite{hendrycks2019benchmarking}, which can be computed by:
\begin{equation}
    \mathrm{CE}_{T}^{f}=\left(\sum_{t=1}^{5} E_{t, T}^{f}\right) \bigg/\left(\sum_{t=1}^{5} E_{t, T}^{ AlexNet}\right),
    \label{CE}
\end{equation}
where $E_{t, T}^{f}$ denotes the error rate of model $f$  on target domain data transformed by corruption $T$ with severity $t$. The $AlexNet$ is trained on clean source domain and tested on corrupted target domain.

The corruption robustness of model $f$ is summarized by averaging Corruption Error values of 15 corruptions introduced in Section~\ref{sec:definition_robustness}: 
$\mathrm{CE}_{\text{Gaussian Noise}}^{f}, \dots , \mathrm{CE}_{\text{Glass Blur}}^{f}$. The results in the mean CE or mCE~\cite{hendrycks2019benchmarking} for short. mCE is calculated by only one setting (\eg, the Ar:Rw setting in Office-Home). For average performance of corruption robustness on the whole dataset (\eg, Office-Home), we need to average mCE values of all settings.

%------------------------------------------------------------------------
\section{Experiments}
%------------------------------------------------------------------------------------
\subsection{Setups}
\paragraph{Datasets.} Office-Home and Office-31 are two benchmark datasets widely adopted for visual domain adaptation algorithms. Experiments are mainly conducted on Office-Home, a relatively challenging dataset. 
% Office is an easier dataset to validate generalization of our methods. 
\textit{Office-Home}~\cite{venkateswara2017officehome} is a challenging medium sized benchmark, which contains 15588 images from 4 domains (Artistic images (Ar), Clip Art (Cl), Product images (Pr), and Real-World images (Rw)). Each domain consists of 65 object classes under daily life environment.
\textit{Office-31}~\cite{saenko2010office31} is a standard benchmark with 4110 images and 31 classes under office environment. There are totally three domains: Amazon (A), Webcam (W) and DSLR (D).

\paragraph{Corruption.} To check the corruption robust of one given DA model, we create the corrupted version of Office-31 and Office-Home by using the corruption types defined by ImageNet-C \cite{hendrycks2019benchmarking}, a widely used benchmark for corruption robustness. For each image, there exists $15$ corruption types with $5$ levels of severity as illustrated in Fig.~\ref{fig:15corruption1}. 
% We further test on these two datasets with spatial transformation and the implementation details can be seen in corresponding experiments.

%------------------------------------------------------------------------------------
\paragraph{Baselines} To illustrate the improvement on CRDA, we apply our method to \textbf{CDAN+TN} \cite{long2018conditional,wang2019TransNorm}, a classic baseline model for domain adaptation. We further apply our method to a SOTA DA model \textbf{DCAN} \cite{Li20DCAN}. 
% In addition, our methods can further maintain or improve models' original accuracy. To validate this, another SOTA model \textbf{SHOT} \cite{liang2020shot} is adopted together with CDAN+TN and DCAN.
Note that the domain discrepancies of CDAN+TN and DCAN are respectively measured by adversarial methods~\cite{goodfellow2014GAN} and MMD~\cite{borgwardt2006integrating}. 
We compare our DDG method with \textbf{AugMix} \cite{hendrycks2019augmix}, a SOTA method for corruption robustness, which aggregates several general transformations such as contrast, equalization and posterization for data augmentation. 
 
 %-----------------------------------------------------------------------------------
 \paragraph{Implementation details}
 For contrastive loss in Equations (\ref{total_loss}), the trade-off $\lambda$ is set to $0.5$. For DDG, we conduct Algorithm \ref{algorithm-DDG} with $\delta=60/255$, $\eta=6$, $n=2$. Network structures and other hyper-parameters are the same as the original DA models.
 
 %-----------------------------------------------------------------------------------
% \subsection{Experiments of Trade-off in TSCL}
% \begin{table*}
%     \centering
%     \caption{Accuracy ($\%$) on clean Office-Home data.}
%     \label{tab:acc_OfficeHome}
%     \small{
%     \setlength{\tabcolsep}{1.0mm}{
%     \begin{tabular}{l|cccccccccccc|c}
%         \toprule
%         Method & Ar$\to$Cl & Ar$\to$Pr & Ar$\to$Rw & Cl$\to$Ar & Cl$\to$Pr & Cl$\to$Rw & Pr$\to$Ar & Pr$\to$Cl & Pr$\to$Rw & Rw$\to$Ar & Rw$\to$Cl & Rw$\to$Pr & Avg \\
%         \hline
%         CDAN+TN &54.1 & 70.3 & 77.9 & 62.2 & 74.4 & 73.7 & 62.4 & 53.1 & 81.0 & 72.8 & 56.9 & 82.5 & 68.4\\
%         DCAN & 57.8 & 76.3 & 82.9 & 68.5 & 72.7 & 76.7 & 68.0 & 56.5 & 82.1 & 73.5 & 60.8 & 83.3 & 71.6\\
%         SHOT & 57.8 & 78.8 & 82.0 & 67.7 & 78.8 & 77.0 & 67.1 & 54.6 & 82.1 & 73.1 & 58.2 & 84.1 & 71.8\\
%         AdvTL & 57.1 & 74.4 & 79.4 & 63.5 & 75.9 & 75.3 & 63.2 & 54.9 & 81.9 & 73.1 & 58.8 & 84.2 & 70.1\\
%         DCAN+AdvTL & 58.1 & 75.2 & 82.9 & 68.7 & 75.0 & 77.6 & 68.1 & 56.6 & 81.8 & 73.9 & 60.6 & 83.1 & 71.8\\
%         TSCL & 55.6 & 71.0 & 79.2 & 63.3 & 75.5 & 75.2 & 62.6 & 54.1 & 81.6 & 73.1 & 57.8 & 83.3 & 69.3\\
%         DCAN+TSCL & 58.1 & 75.6 & 82.8 & 69.0 & 74.7 & 77.0 & 68.4 & 57.2 & 82.3 & 74.2 & 61.2 & 83.4 & 72.0\\
%         SHOT+TSCL & 58.0 & 79.0 & 82.4 & 68.2 & 79.9 & 78.4 & 67.1 & 55.0 & 81.7 & 73.4 & 58.1 & 84.5 & 72.1\\
%         \bottomrule
%     \end{tabular}
%     }
%     }
% \end{table*}
\begin{table*}
    \centering
    \caption{Accuracy ($\%$) on clean Office-Home data (ResNet-50).}
    \label{tab:acc_OfficeHome}
    \resizebox{140mm}{!}{%\small{
    \setlength{\tabcolsep}{0.8mm}{
    \begin{tabular}{l|cccccccccccc|c}
        \toprule
        Method & Ar$\to$Cl & Ar$\to$Pr & Ar$\to$Rw & Cl$\to$Ar & Cl$\to$Pr & Cl$\to$Rw & Pr$\to$Ar & Pr$\to$Cl & Pr$\to$Rw & Rw$\to$Ar & Rw$\to$Cl & Rw$\to$Pr & Avg ($\uparrow$) \\
        \hline
        ResNet & 34.9 & 50.0 & 58.0 & 37.4 & 41.9 & 46.2 & 38.5 & 31.2 & 60.4 & 53.9 & 41.2 & 59.9 & 46.1\\
        DAN & 43.6 & 57.0 & 67.9 & 45.8 & 56.5 & 60.4 & 44.0 & 43.6 & 67.7 & 63.1 & 51.5 & 74.3 & 56.3\\
        DANN & 45.6 & 59.3 & 70.1 & 47.0 & 58.5 & 60.9 & 46.1 & 43.7 & 68.5 & 63.2 & 51.8 & 76.8 & 57.6\\
        JAN & 45.9 & 61.2 & 68.9 & 50.4 & 59.7 & 61.0 & 45.8 & 43.4 & 70.3 & 63.9 & 52.4 & 76.8 & 58.3\\
        DWT & 50.3 & 72.1 & 77.0 & 59.6 & 69.3 & 70.2 & 58.3 & 48.1 & 77.3 & 69.3 & 53.6 & 82.0 & 65.6\\
        CDAN & 50.7 & 70.6 & 76.0 & 57.6 & 70.0 & 70.0 & 57.4 & 50.9 & 77.3 & 70.9 & 56.7 & 81.6 & 65.8\\
        TADA & 53.1 & 72.3 & 77.2 & 59.1 & 71.2 & 72.1 & 59.7 & 53.1 & 78.4 & 72.4 & 60.0 & 82.9 & 67.6\\
        SymNets & 47.7 & 72.9 & 78.5 & 64.2 & 71.3 & 74.2 & 64.2 & 48.8 & 79.5 & 74.5 & 52.6 & 82.7 & 67.6\\
        MDD & 54.9 & 73.7 & 77.8 & 60.0 & 71.4 & 71.8 & 61.2 & 53.6 & 78.1 & 72.5 & 60.2 & 82.3 & 68.1\\
        \hline
        CDAN+TN & 54.1 & 70.3 & 77.9 & 62.2 & 74.4 & 73.7 & 62.4 & 53.1 & 81.0 & 72.8 & 56.9 & 82.5 & 68.4\\
        +AugMix & 50.5 & 70.1 & 74.9 & 56.7 & 69.6 & 68.7 & 53.9 & 50.4 & 75.9 & 67.9 & 58.2 & 80.6 & 64.8\\
        % +TSCL & 55.6 & 71.0 & 79.2 & 63.3 & 75.5 & 75.2 & 62.6 & 54.1 & 81.6 & 73.1 & 57.8 & 83.3 & 69.3\\
        \textbf{+DDG} & 57.1 & 74.4 & 79.4 & 63.5 & 75.9 & 75.3 & 63.2 & 54.9 & 81.9 & 73.1 & 58.8 & 84.2 & \textbf{70.1}\\
        \hline
        DCAN & 57.8 & \textbf{76.3} & 82.9 & 68.5 & 72.7 & 76.7 & 68.0 & 56.5 & 82.1 & 73.5 & \textbf{60.8} & \textbf{83.3} & 71.6\\
        +AugMix & 55.8 & 74.8 & 82.5 & 67.6 & 72.1 & 76.1 & 67.5 & 55.0 & \textbf{82.5} & 73.4 & 59.0 & 82.5 & 70.7\\
        % +TSCL & 58.1 & 75.6 & 82.8 & 69.0 & 74.7 & 77.0 & 68.4 & 57.2 & 82.3 & 74.2 & 61.2 & 83.4 & 72.0\\
        \textbf{+DDG} & \textbf{58.1} & 75.2 & \textbf{82.9} & \textbf{68.7} & \textbf{75.0} & \textbf{77.6} & \textbf{68.1} & \textbf{56.6} & 81.8 & \textbf{73.9} & 60.6 & 83.1 & \textbf{71.8}\\
        % \hline
        % SHOT & 57.8 & 78.8 & 82.0 & 67.7 & 78.8 & 77.0 & 67.1 & 54.6 & 82.1 & 73.1 & 58.2 & 84.1 & 71.8\\
        % +TSCL & 58.0 & 79.0 & 82.4 & 68.2 & 79.9 & 78.4 & 67.1 & 55.0 & 81.7 & 73.4 & 58.1 & 84.5 & 72.1\\
        \bottomrule
    \end{tabular}
    }
    }
    \vspace{-1.5mm}
\end{table*}

% % In this section, we evaluate our TSCL framework method for stage 2-(a). Table \ref{tab:err_TSCL_home} precisely reports the error rate on clean data and some corruptions with the most severe severity. The full results can be seen in SM. Results show that our TSCL can significantly improve the model’s robustness against a specific corruption. Furthermore, TSCL can further improves the model’s original performance instead of negative effect brought by pure data augmentation. In another word, TSCL does not conduct trade-off over robustness and accuracy. 
% Assuming testing corruptions available while training, we precisely report the error rate of TSCL on clean data and some corruptions with the most severe severity in Table \ref{tab:err_TSCL_home}. The full results can be seen in SM. Results show that our TSCL can significantly improve the model’s robustness against a specific known corruption. Furthermore, instead of the negative effect brought by simultaneous learning of pure data augmentation, TSCL further improves the model’s average clean accuracy that reflects domain invariance. In another word, the step-by-step learning scheme of TSCL balances the trade-off between domain invariance and corruption robustness well. 

% 

\begin{figure*}
    \centering
    \includegraphics[width=0.92\textwidth,height=5.4cm]{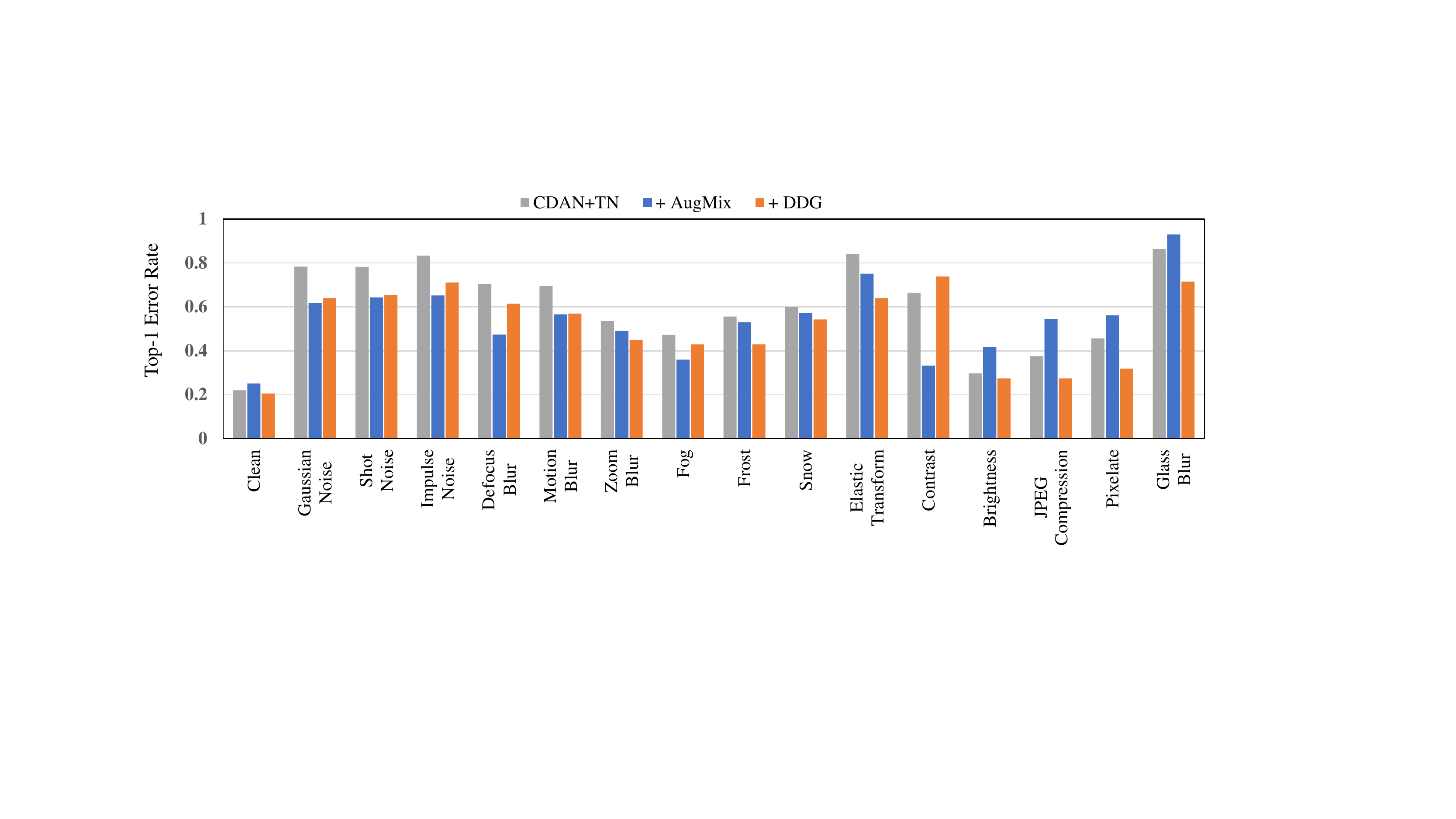}
    \caption{Error rate ($\%$) on Ar$\to$Rw of Office-Home for different corruptions with the most severe level under CRDA (ResNet-50).}
    \label{fig:exp3}
    \vspace{-3mm}
\end{figure*}

%-----------------------------------------------------------------------------------
\subsection{Experiments in CRDA}
We compare DDG with AugMix and the original DA models on both corruption robustness and original performance. In addition, we further set an empirical lower bound for CRDA calculated by simply replacing DDG-generated samples in Algorithm \ref{algorithm-DDG} with corresponding corrupted samples,which means models that reach the lower bound can be robust against unpredictable corruptions as if they are already known beforehand.
 
Fig.~\ref{fig:exp3} shows the error rate on clean data and 15 different corruptions with the most severe severity in a single setting. It is reported that DDG achieves improvement over most kinds of corruptions with about 10 percentages in average than the original model and 3 percentages than AugMix.  We observe that AugMix only achieves relatively high robustness on several corruptions such as Fog and Contrast. The reason possibly is that these corruptions can be implied by the aggregation of the general transformations defined by AugMix, however, corruptions like Pixelate may lie out of the implied set. Instead, our DDG achieves a much more generalizable implied set thanks to reasonable utilization of domain discrepancy information. It is worthy of noting that DDG can also improve the original accuracy, which is detailedly reported in Table \ref{tab:acc_OfficeHome}.

% \begin{table*}
%     \centering
%     \caption{mCE on all settings of Office-Home dataset for stage2-(b).}
%     \label{tab:mCE_OfficeHome}
%     \small{
%     \setlength{\tabcolsep}{1.1mm}{
%     \begin{tabular}{l|cccccccccccc|c}
%         \toprule
%         Method & Ar$\to$Cl & Ar$\to$Pr & Ar$\to$Rw & Cl$\to$Ar & Cl$\to$Pr & Cl$\to$Rw & Pr$\to$Ar & Pr$\to$Cl & Pr$\to$Rw & Rw$\to$Ar & Rw$\to$Cl & Rw$\to$Pr & Avg \\
%         \hline
%         Ori & 68.6 & 60.5 & 57.2 & 64.1 & 57.4 & 56.2 & 64.0 & 69.8 & 52.8 & 61.3 & 69.5 & 58.2 &  \\
%         Ori+AugMix & 71.9 & 65.9 & 61.7 & 74.0 & 68.4 & 66.7 & 75.0 & 78.0 & 64.1 & 68.5 & 73.8 & 64.2 &  \\
%         Ori+AdvTL & 60.1 & 51.5 & 51.1 & 65.0 & 60.1 & 57.2 & 60.5 & 63.3 & 51.9 & 61.7 & 66.2 & 50.0 &  \\
%         \bottomrule
%     \end{tabular}
%     }
%     }
% \end{table*}

\begin{table}
    \centering
    \caption{mCE ($\%$) on all cross-domain settings of Office-Home dataset under CRDA (ResNet-50).}
    \label{tab:mCE_OfficeHome}
    \setlength{\tabcolsep}{1.2mm}{
    \small{
    \begin{tabular}{c|ccc|ccc|c}
        \toprule
        \multirow{2}{*}{Settings} & \multicolumn{3}{c|}{CDAN+TN} &\multicolumn{3}{c|}{DCAN} & Lower \\
        \cline{2-7}
         & - & AugMix & Ours & - & AugMix & Ours & Bound \\
        \hline
        Ar$\to$Cl & 69.7 & 69.3 & 59.8 & 61.4 & 63.2 & 58.3 & 52.2 \\
        Ar$\to$Pr & 62.2 & 52.8 & 52.2 & 51.0 & 49.3 & 48.0 & 37.7 \\
        Ar$\to$Rw & 59.2 & 56.0 & 48.0 & 47.9 & 46.0 & 44.6 & 30.8 \\
        Cl$\to$Ar & 65.4 & 73.7 & 66.0 & 58.6 & 57.4 & 55.2 & 45.8 \\
        Cl$\to$Pr & 58.7 & 56.6 & 51.9 & 58.0 & 57.3 & 51.6 & 34.6 \\
        Cl$\to$Rw & 57.6 & 61.3 & 56.2 & 52.0 & 53.8 & 47.5 & 34.7 \\
        Pr$\to$Ar & 65.5 & 72.2 & 63.6 & 61.1 & 58.9 & 55.1 & 44.7 \\
        Pr$\to$Cl & 70.9 & 71.4 & 63.6 & 66.2 & 65.3 & 62.2 & 54.9 \\
        Pr$\to$Rw & 55.0 & 55.3 & 48.8 & 53.7 & 48.5 & 49.4 & 25.7 \\
        Rw$\to$Ar & 63.2 & 62.4 & 57.8 & 63.0 & 59.2 & 57.7 & 36.0 \\
        Rw$\to$Cl & 70.4 & 61.4 & 64.4 & 64.0 & 64.2 & 59.9 & 52.9 \\
        Rw$\to$Pr & 60.9 & 52.6 & 47.9 & 58.4 & 52.5 & 52.6 & 26.8 \\
        \hline
        Avg ($\downarrow$) & 63.2 & 62.1 & \textbf{56.7} & 57.9 & 56.3 & \textbf{53.5} & 39.7 \\
        \bottomrule
    \end{tabular}
    }
    }
    \vspace{-1.5mm}
\end{table}

% \begin{table}
%     \centering
%     \caption{mCE ($\%$) on all cross-domain settings of Office-31 dataset under CRDA.}
%     \label{tab:mCE_Office}
%     \setlength{\tabcolsep}{2.4mm}{
%     \small{
%     \begin{tabular}{c|ccc|c}
%         \toprule
%         \multirow{2}{*}{Settings} & \multicolumn{3}{c|}{CDAN+TN} & Lower \\
%         \cline{2-4}
%          & - & AugMix & Ours & Bould \\
%         \hline
%         A$\to$D & 60.3 & 46.0 & 48.9 & 12.3\\
%         A$\to$W & 59.6 & 65.2 & 46.9 & 10.1\\
%         D$\to$A & 64.1 & 66.0 & 62.4 & 33.8\\
%         D$\to$W & 83.5 & 59.9 & 69.7 & 4.6\\
%         W$\to$A & 64.7 & 67.9 & 60.1 & 36.4\\
%         W$\to$D & 120.2 & 64.3 & 70.9 & 0.7\\
%         \hline
%         Avg ($\downarrow$) & 75.4 & 61.6 & \textbf{59.8} & 16.3\\
%         \bottomrule
%     \end{tabular}
%     }
%     }
% \end{table}
\begin{table}
    \centering
    \caption{mCE ($\%$) on all cross-domain settings of Office-31 dataset under CRDA (ResNet-50).}
    \label{tab:mCE_Office}
    \setlength{\tabcolsep}{1.15mm}{
    \small{
    \begin{tabular}{c|ccc|ccc|c}
        \toprule
        \multirow{2}{*}{Settings} & \multicolumn{3}{c|}{CDAN+TN} &\multicolumn{3}{c|}{DCAN} & Lower \\
        \cline{2-7}
         & - & AugMix & Ours & - & AugMix & Ours & Bound \\
        \hline
        A$\to$D & 60.3 & 46.0 & 48.9 & 42.3 & 40.5 & 36.9 & 12.3\\
        A$\to$W & 59.6 & 65.2 & 46.9 & 40.0 & 32.4 & 28.6 & 10.1\\
        D$\to$A & 64.1 & 66.0 & 62.4 & 50.3 & 53.0 & 46.6 & 33.8\\
        D$\to$W & 83.5 & 59.9 & 69.7 & 45.2 & 48.1 & 35.9 & 4.6\\
        W$\to$A & 64.7 & 67.9 & 60.1 & 60.1 & 59.8 & 55.0 & 36.4\\
        W$\to$D & 120.2 & 64.3 & 70.9 & 46.6 & 54.6 & 44.5 & 0.7\\
        \hline
        Avg ($\downarrow$) & 75.4 & 61.6 & \textbf{59.8} & 47.4 & 48.1 & \textbf{41.3} & 16.3 \\
        \bottomrule
    \end{tabular}
    }
    }
    \vspace{-3mm}
\end{table}
 
Fig.~\ref{fig:exp3} also shows that DDG gains no improvement over the Contrast corruption. We argue the reason probably is that this corruption conducts too much shift in sample space which goes beyond the scope of Assumption~\ref{assum:assum1}. Possible solutions include loosening the constraint in Assumption~\ref{assum:assum1} and considering the invariant order of pixel values (Contrast does not change the order of pixel values). We leave them for future work.

We report the overall robustness under the whole Office-Home and Office-31 datasets in Table \ref{tab:mCE_OfficeHome} and Table \ref{tab:mCE_Office} with the standard robustness metric mCE. Results show that even the SOTA model AugMix has sub-optimal results or even worse  performance in some settings, which indicates the challenge of CRDA. Despite the challenging task, DDG still gains improvement with obvious margins. Meanwhile, there is still a long way for further study to reach the lower bound in CRDA.
Besides robustness, DDG also maintains or improves the original accuracy of DA models. Results are shown in Table \ref{tab:acc_OfficeHome}. 

\subsection{Empirical analysis}
\label{sec:empirical}

\paragraph{Ablation.}
\begin{table}[t]
    \centering
    \caption{mCE ($\%$) on Ar$\to$Rw for different updata strides $\eta$ in DDG}
    \label{tab:Ablation-delta}
    \small{
    \setlength{\tabcolsep}{5mm}{
    \begin{tabular}{ccc|c}
        \toprule
        $\delta$&$\eta$&$n$&mCE ($\downarrow$)\\
        \hline
        $60/255$ & $> 2\delta$ & 2 & 48.0\\ 
        $60/255$ & $> 2\delta$ & 10 & 49.6\\ 
        $60/255$ & $15/255$ & 10 & 51.6\\ 
        $60/255$ & $6/255$ & 10 & 52.8\\ 
        \bottomrule
    \end{tabular}
    }
    }
    \vspace{-3mm}
\end{table}

In Section~\ref{sec:theory}, we theoretically show only considering the edge points on the $\delta$ cycle is enough for DDG, which is realized by setting stride $\eta > 2\delta$ in Algorithm \ref{algorithm-DDG}. Table \ref{tab:Ablation-delta} reports the mCE under CRDA with different strides $\eta$. Results show that the setting $\eta > 2\delta$, which results in only considering the edge points, gains best performance. We conclude the reason why settings $\eta \ll \delta$ perform relatively poorer is that they may wander in the $\delta$ cycle instead of reaching the large-shift points on the edge, which causes a lack of consideration on large-shift corruptions like contrast and blurs. Table \ref{tab:Ablation-delta} also shows that only 2 update steps is enough for DDG to gain corruption robustness.
More ablation results can be seen in Appendix.
\begin{figure}
    \subfloat[Cosine distance ($\uparrow$)]{
    \begin{minipage}[c][0.4\textwidth]{0.45\textwidth}
    \centering
    \includegraphics[width=\linewidth]{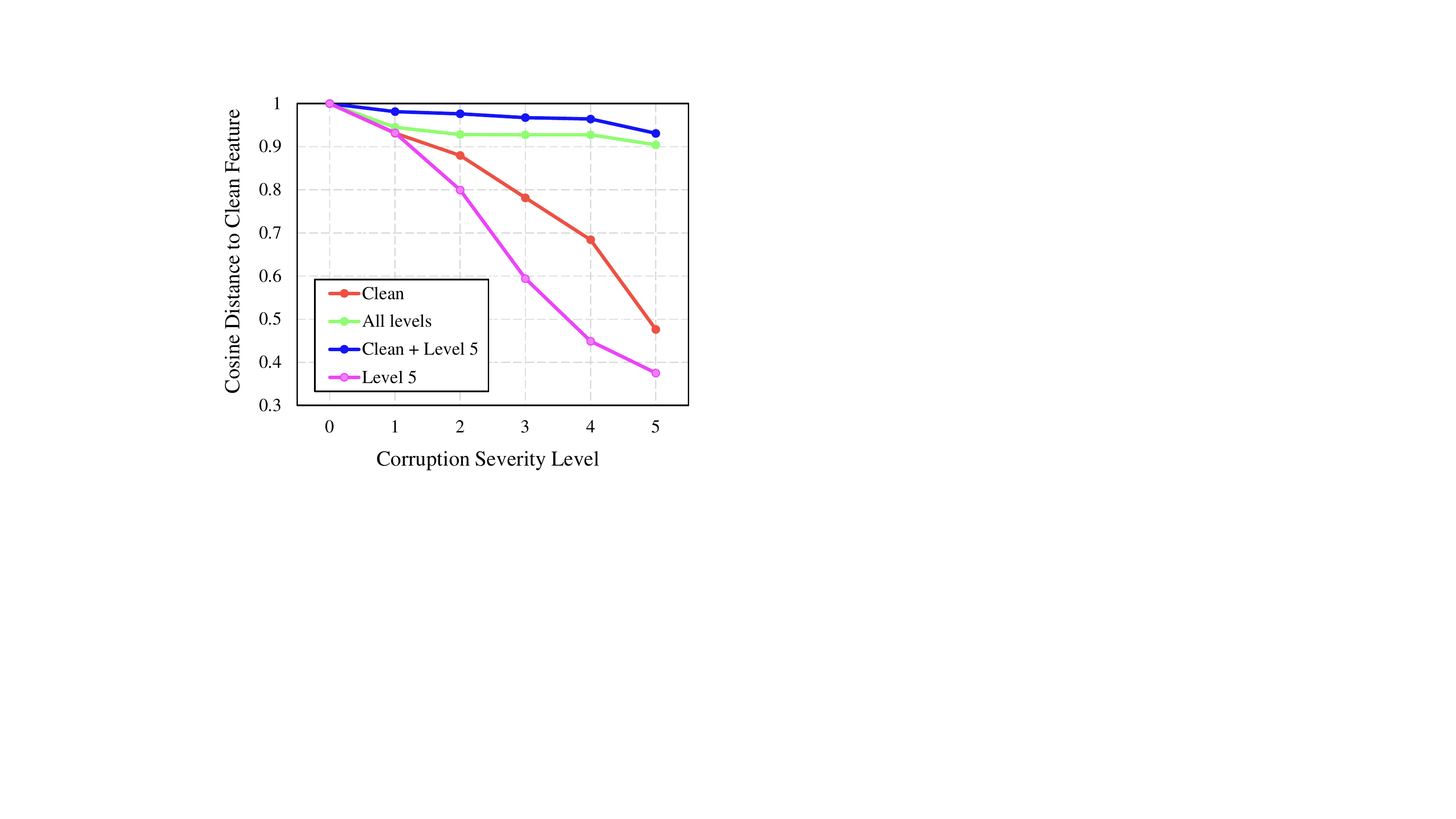}
    \end{minipage}}
    % 	\hfill
    \subfloat[Euclidean distance ($\downarrow$)]{
    \begin{minipage}[c][0.4\textwidth]{0.45\textwidth}
    \centering
    \includegraphics[width=\linewidth]{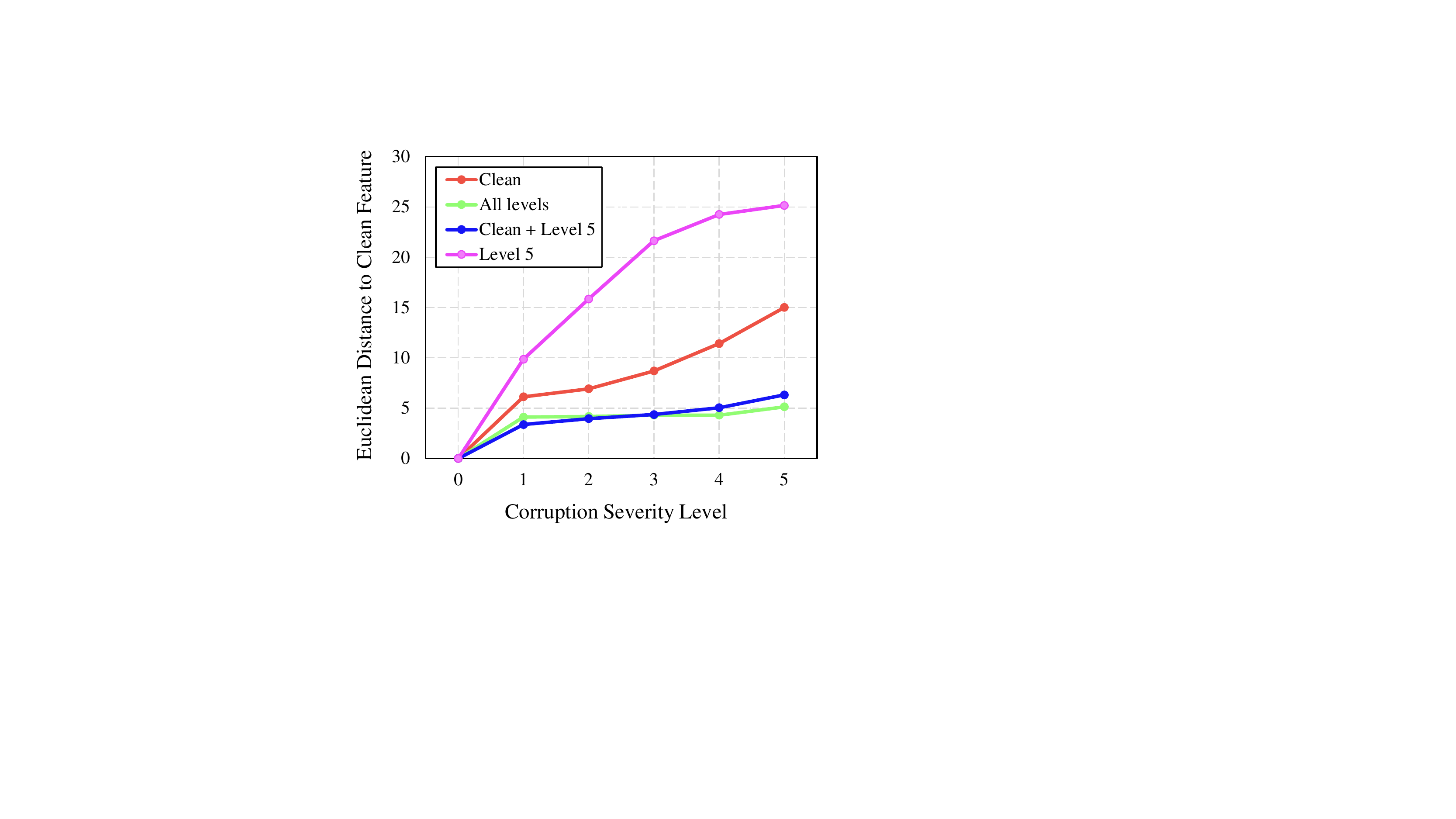}
    \end{minipage}}
    \caption{Feature distance between a clean image and the versions corrupted by five severity levels of Gaussian Noise on the basis of different models: \textbf{Clean} (a ResNet-50 trained on clean data), \textbf{All levels} (trained on all severity levels of Gaussian Noise), \textbf{Clean + Level 5} (trained on clean data and severity level 5 Gaussian Noise), \textbf{Level 5} (trained on data only corrupted by severity level 5).}
    \vspace{-3mm}
    \label{fig:order}
\end{figure}

\paragraph{Teacher-student warm-up scheme.} 
In this section, we aim to evaluate the effectiveness of the teacher-student warm-up scheme. To compare with simply augmenting samples with testing corruptions, we replace DDG-generated samples in Algorithm \ref{algorithm-DDG} with corresponding testing corruptions. Table \ref{tab:err_TSCL_home} precisely reports the error rate on original data and corruptions with the most severe severity.  Results show that our warm-up scheme can significantly improve the model’s robustness against a specific corruption. Furthermore, our scheme can further improves the model’s original performance instead of negative effect brought by pure data augmentation. In another word, the teacher-student warm-up scheme does not conduct trade-off over robustness and accuracy. 
\begin{table}
    \centering
    \caption{Error rate ($\%$) under a specific known corruption on the Ar$\to$Rw setting of Office-Home dataset for teacher-student warm-up shceme.}
    \label{tab:err_TSCL_home}
    \setlength{\tabcolsep}{1.6mm}{
    \small{
    \begin{tabular}{c|ccc|c|ccc}
        \toprule
        \multirow{2}{*}{Corruption} & \multicolumn{3}{c|}{CDAN+TN} &\multirow{2}{*}{Corruption} & \multicolumn{3}{c}{CDAN+TN} \\
        \cline{2-4}
        \cline{6-8}
         & -  & DataAug & Ours  & & - & DataAug & Ours\\
        \hline
        Clean & 22.1 & 24.6 & 21.2 & Frost & 55.7 & 30.6 & 25.5\\
        Gaussian Noise & 78.4 & 33.4 & 27.8 & Snow & 60.0 & 31.2 & 23.9\\
        Shot Noise & 78.3 & 34.9 & 26.9 & Elastic Transform & 84.2 & 30.0 & 23.3\\
        Impulse Noise & 83.4 & 33.4 & 24.4 & Contrast & 66.4 & 24.6 & 23.0\\
        Defocus Blur & 70.4 & 34.9 & 26.7 & Brightness & 29.7 & 26.5 & 23.6\\
        Motion Blur & 69.5 & 28.4 & 24.3 & JPEG Compression & 37.6 & 29.0 & 23.2\\
        Zoom Blur & 53.6 & 29.4 & 23.1 & Pixelate & 45.7 & 26.9 & 22.6\\
        Fog & 47.3 & 24.6 & 23.6 & Glass Blur & 86.4 & 37.0 & 27.8\\
        \bottomrule
    \end{tabular}
    }
    }
    \vspace{-1.5mm}
\end{table}

\paragraph{Order-invariant representations of corruptions.}
We empirically observe that the distance between the corrupted versions and the corresponding clean image in feature space (\textbf{feature distance} for precise) is always positively correlated to the severity of corruptions. Training details are shown in the supplementary material. Fig.~\ref{fig:order} shows that even only trained on clean and corrupted samples with a single severity (Clean + Level 5), models still learn the positive correlation instead of relatively nearer feature distance on a specific severity. In a word, networks always learn order-invariant representations in feature space for different levels of severity, which contributes to Assumption~\ref{assum:assum2} in Section~\ref{sec:wc}.

% \input{latex/tabs/acc_rawOffice}

%------------------------------------------------------------------------
\section{Conclusion}
In this paper, we throw a new sight to domain adaptation to investigate a more realistic new task, Corruption-agnostic Robust Domain Adaptation (CRDA). Taking domain information into consideration, we present a new idea for corruption robustness called Domain Discrepancy Generator (DDG) that mimic unpredictable corruptions via generating samples most enlarging domain discrepancy. Besides, we propose a teacher-student warm-up scheme via contrastive loss to enhance the constraints on unlabelled target domains and stabilize the early training stage feature. Empirical results justify that DDG outperforms existing baselines on original accuracy and achieves better corruption robustness.

%----------------------------------------------------------

\bibliographystyle{ACM-Reference-Format}
\bibliography{main}


\begin{thebibliography}{46}

%%% the Web address.

\ifx \showCODEN    \undefined \def \showCODEN     #1{\unskip}     \fi
\ifx \showDOI      \undefined \def \showDOI       #1{#1}\fi
\ifx \showISBNx    \undefined \def \showISBNx     #1{\unskip}     \fi
\ifx \showISBNxiii \undefined \def \showISBNxiii  #1{\unskip}     \fi
\ifx \showISSN     \undefined \def \showISSN      #1{\unskip}     \fi
\ifx \showLCCN     \undefined \def \showLCCN      #1{\unskip}     \fi
\ifx \shownote     \undefined \def \shownote      #1{#1}          \fi
\ifx \showarticletitle \undefined \def \showarticletitle #1{#1}   \fi
\ifx \showURL      \undefined \def \showURL       {\relax}        \fi
% invisible to TeX
\providecommand\bibfield[2]{#2}
\providecommand\bibinfo[2]{#2}
\providecommand\natexlab[1]{#1}
\providecommand\showeprint[2][]{arXiv:#2}

\bibitem[\protect\citeauthoryear{Borgwardt, Gretton, Rasch, Kriegel,
  Sch{\"o}lkopf, and Smola}{Borgwardt et~al\mbox{.}}{2006}]%
        {borgwardt2006integrating}
\bibfield{author}{\bibinfo{person}{Karsten~M Borgwardt},
  \bibinfo{person}{Arthur Gretton}, \bibinfo{person}{Malte~J Rasch},
  \bibinfo{person}{Hans-Peter Kriegel}, \bibinfo{person}{Bernhard
  Sch{\"o}lkopf}, {and} \bibinfo{person}{Alex~J Smola}.}
  \bibinfo{year}{2006}\natexlab{}.
\newblock \showarticletitle{Integrating structured biological data by kernel
  maximum mean discrepancy}.
\newblock \bibinfo{journal}{\emph{Bioinformatics}} \bibinfo{volume}{22},
  \bibinfo{number}{14} (\bibinfo{year}{2006}), \bibinfo{pages}{e49--e57}.
\newblock


\bibitem[\protect\citeauthoryear{Chen, Chen, Jiang, and Jin}{Chen
  et~al\mbox{.}}{2019}]%
        {chen2019joint}
\bibfield{author}{\bibinfo{person}{Chao Chen}, \bibinfo{person}{Zhihong Chen},
  \bibinfo{person}{Boyuan Jiang}, {and} \bibinfo{person}{Xinyu Jin}.}
  \bibinfo{year}{2019}\natexlab{}.
\newblock \showarticletitle{Joint domain alignment and discriminative feature
  learning for unsupervised deep domain adaptation}. In
  \bibinfo{booktitle}{\emph{Proceedings of the AAAI Conference on Artificial
  Intelligence}}, Vol.~\bibinfo{volume}{33}. \bibinfo{pages}{3296--3303}.
\newblock


\bibitem[\protect\citeauthoryear{Chen, Kornblith, Norouzi, and Hinton}{Chen
  et~al\mbox{.}}{2020a}]%
        {chen2020simple}
\bibfield{author}{\bibinfo{person}{Ting Chen}, \bibinfo{person}{Simon
  Kornblith}, \bibinfo{person}{Mohammad Norouzi}, {and}
  \bibinfo{person}{Geoffrey Hinton}.} \bibinfo{year}{2020}\natexlab{a}.
\newblock \showarticletitle{A simple framework for contrastive learning of
  visual representations}. In \bibinfo{booktitle}{\emph{ICML}}.
\newblock


\bibitem[\protect\citeauthoryear{Chen, Liu, Chang, Cheng, Amini, and Wang}{Chen
  et~al\mbox{.}}{2020b}]%
        {chen2020adversarial}
\bibfield{author}{\bibinfo{person}{Tianlong Chen}, \bibinfo{person}{Sijia Liu},
  \bibinfo{person}{Shiyu Chang}, \bibinfo{person}{Yu Cheng},
  \bibinfo{person}{Lisa Amini}, {and} \bibinfo{person}{Zhangyang Wang}.}
  \bibinfo{year}{2020}\natexlab{b}.
\newblock \showarticletitle{Adversarial Robustness: From Self-Supervised
  Pre-Training to Fine-Tuning}. In \bibinfo{booktitle}{\emph{Proceedings of the
  IEEE/CVF Conference on Computer Vision and Pattern Recognition}}.
  \bibinfo{pages}{699--708}.
\newblock


\bibitem[\protect\citeauthoryear{Chen, Li, Sakaridis, Dai, and Van~Gool}{Chen
  et~al\mbox{.}}{2018}]%
        {chen2018domain}
\bibfield{author}{\bibinfo{person}{Yuhua Chen}, \bibinfo{person}{Wen Li},
  \bibinfo{person}{Christos Sakaridis}, \bibinfo{person}{Dengxin Dai}, {and}
  \bibinfo{person}{Luc Van~Gool}.} \bibinfo{year}{2018}\natexlab{}.
\newblock \showarticletitle{Domain adaptive faster r-cnn for object detection
  in the wild}. In \bibinfo{booktitle}{\emph{Proceedings of the IEEE conference
  on computer vision and pattern recognition}}. \bibinfo{pages}{3339--3348}.
\newblock


\bibitem[\protect\citeauthoryear{Cubuk, Zoph, Shlens, and Le}{Cubuk
  et~al\mbox{.}}{2020}]%
        {cubuk2020randaugment}
\bibfield{author}{\bibinfo{person}{Ekin~D Cubuk}, \bibinfo{person}{Barret
  Zoph}, \bibinfo{person}{Jonathon Shlens}, {and} \bibinfo{person}{Quoc~V Le}.}
  \bibinfo{year}{2020}\natexlab{}.
\newblock \showarticletitle{Randaugment: Practical automated data augmentation
  with a reduced search space}. In \bibinfo{booktitle}{\emph{Proceedings of the
  IEEE/CVF Conference on Computer Vision and Pattern Recognition Workshops}}.
  \bibinfo{pages}{702--703}.
\newblock


\bibitem[\protect\citeauthoryear{Dodge and Karam}{Dodge and Karam}{2017a}]%
        {dodge2017quality}
\bibfield{author}{\bibinfo{person}{Samuel Dodge} {and} \bibinfo{person}{Lina
  Karam}.} \bibinfo{year}{2017}\natexlab{a}.
\newblock \showarticletitle{Quality resilient deep neural networks}.
\newblock \bibinfo{journal}{\emph{arXiv preprint arXiv:1703.08119}}
  (\bibinfo{year}{2017}).
\newblock


\bibitem[\protect\citeauthoryear{Dodge and Karam}{Dodge and Karam}{2017b}]%
        {dodge2017study}
\bibfield{author}{\bibinfo{person}{Samuel Dodge} {and} \bibinfo{person}{Lina
  Karam}.} \bibinfo{year}{2017}\natexlab{b}.
\newblock \showarticletitle{A study and comparison of human and deep learning
  recognition performance under visual distortions}. In
  \bibinfo{booktitle}{\emph{2017 26th international conference on computer
  communication and networks (ICCCN)}}. IEEE, \bibinfo{pages}{1--7}.
\newblock


\bibitem[\protect\citeauthoryear{Ganin, Ustinova, Ajakan, Germain, Larochelle,
  Laviolette, Marchand, and Lempitsky}{Ganin et~al\mbox{.}}{2016}]%
        {ganin2016DANN}
\bibfield{author}{\bibinfo{person}{Yaroslav Ganin}, \bibinfo{person}{Evgeniya
  Ustinova}, \bibinfo{person}{Hana Ajakan}, \bibinfo{person}{Pascal Germain},
  \bibinfo{person}{Hugo Larochelle}, \bibinfo{person}{Fran{\c{c}}ois
  Laviolette}, \bibinfo{person}{Mario Marchand}, {and} \bibinfo{person}{Victor
  Lempitsky}.} \bibinfo{year}{2016}\natexlab{}.
\newblock \showarticletitle{Domain-adversarial training of neural networks}.
\newblock \bibinfo{journal}{\emph{The Journal of Machine Learning Research}}
  \bibinfo{volume}{17}, \bibinfo{number}{1} (\bibinfo{year}{2016}),
  \bibinfo{pages}{2096--2030}.
\newblock


\bibitem[\protect\citeauthoryear{Geirhos, Temme, Rauber, Sch{\"u}tt, Bethge,
  and Wichmann}{Geirhos et~al\mbox{.}}{2018}]%
        {geirhos2018generalisation}
\bibfield{author}{\bibinfo{person}{Robert Geirhos}, \bibinfo{person}{Carlos~RM
  Temme}, \bibinfo{person}{Jonas Rauber}, \bibinfo{person}{Heiko~H Sch{\"u}tt},
  \bibinfo{person}{Matthias Bethge}, {and} \bibinfo{person}{Felix~A Wichmann}.}
  \bibinfo{year}{2018}\natexlab{}.
\newblock \showarticletitle{Generalisation in humans and deep neural networks}.
  In \bibinfo{booktitle}{\emph{Advances in neural information processing
  systems}}. \bibinfo{pages}{7538--7550}.
\newblock


\bibitem[\protect\citeauthoryear{Goodfellow, Pouget-Abadie, Mirza, Xu,
  Warde-Farley, Ozair, Courville, and Bengio}{Goodfellow et~al\mbox{.}}{2014}]%
        {goodfellow2014GAN}
\bibfield{author}{\bibinfo{person}{Ian Goodfellow}, \bibinfo{person}{Jean
  Pouget-Abadie}, \bibinfo{person}{Mehdi Mirza}, \bibinfo{person}{Bing Xu},
  \bibinfo{person}{David Warde-Farley}, \bibinfo{person}{Sherjil Ozair},
  \bibinfo{person}{Aaron Courville}, {and} \bibinfo{person}{Yoshua Bengio}.}
  \bibinfo{year}{2014}\natexlab{}.
\newblock \showarticletitle{Generative adversarial nets}. In
  \bibinfo{booktitle}{\emph{Advances in Neural Information Processing Systems
  (NIPS)}}. \bibinfo{pages}{2672--2680}.
\newblock


\bibitem[\protect\citeauthoryear{Gopalan, Li, and Chellappa}{Gopalan
  et~al\mbox{.}}{2011}]%
        {gopalan2011domain}
\bibfield{author}{\bibinfo{person}{Raghuraman Gopalan}, \bibinfo{person}{Ruonan
  Li}, {and} \bibinfo{person}{Rama Chellappa}.}
  \bibinfo{year}{2011}\natexlab{}.
\newblock \showarticletitle{Domain adaptation for object recognition: An
  unsupervised approach}. In \bibinfo{booktitle}{\emph{2011 international
  conference on computer vision}}. IEEE, \bibinfo{pages}{999--1006}.
\newblock


\bibitem[\protect\citeauthoryear{Han, Gui, Cui, and Yin}{Han
  et~al\mbox{.}}{2020}]%
        {han2020towards}
\bibfield{author}{\bibinfo{person}{Zhongyi Han}, \bibinfo{person}{Xian-Jin
  Gui}, \bibinfo{person}{Chaoran Cui}, {and} \bibinfo{person}{Yilong Yin}.}
  \bibinfo{year}{2020}\natexlab{}.
\newblock \showarticletitle{Towards Accurate and Robust Domain Adaptation under
  Noisy Environments}. In \bibinfo{booktitle}{\emph{Proceedings of the
  Twenty-Ninth International Joint Conference on Artificial Intelligence,
  {IJCAI-20}}}, \bibfield{editor}{\bibinfo{person}{Christian Bessiere}} (Ed.).
  \bibinfo{publisher}{IJCAI}, \bibinfo{pages}{2269--2276}.
\newblock
\newblock
\shownote{Main track.}


\bibitem[\protect\citeauthoryear{Hendrycks, Basart, Mu, Kadavath, Wang,
  Dorundo, Desai, Zhu, Parajuli, Guo, Song, Steinhardt, and Gilmer}{Hendrycks
  et~al\mbox{.}}{2020a}]%
        {hendrycks2020many}
\bibfield{author}{\bibinfo{person}{Dan Hendrycks}, \bibinfo{person}{Steven
  Basart}, \bibinfo{person}{Norman Mu}, \bibinfo{person}{Saurav Kadavath},
  \bibinfo{person}{Frank Wang}, \bibinfo{person}{Evan Dorundo},
  \bibinfo{person}{Rahul Desai}, \bibinfo{person}{Tyler Zhu},
  \bibinfo{person}{Samyak Parajuli}, \bibinfo{person}{Mike Guo},
  \bibinfo{person}{Dawn Song}, \bibinfo{person}{Jacob Steinhardt}, {and}
  \bibinfo{person}{Justin Gilmer}.} \bibinfo{year}{2020}\natexlab{a}.
\newblock \showarticletitle{The Many Faces of Robustness: A Critical Analysis
  of Out-of-Distribution Generalization}.
\newblock \bibinfo{journal}{\emph{arXiv preprint arXiv:2006.16241}}
  (\bibinfo{year}{2020}).
\newblock


\bibitem[\protect\citeauthoryear{Hendrycks and Dietterich}{Hendrycks and
  Dietterich}{2019}]%
        {hendrycks2019benchmarking}
\bibfield{author}{\bibinfo{person}{Dan Hendrycks} {and} \bibinfo{person}{Thomas
  Dietterich}.} \bibinfo{year}{2019}\natexlab{}.
\newblock \showarticletitle{Benchmarking neural network robustness to common
  corruptions and perturbations}. In \bibinfo{booktitle}{\emph{ICLR}}.
\newblock


\bibitem[\protect\citeauthoryear{Hendrycks, Mazeika, Kadavath, and
  Song}{Hendrycks et~al\mbox{.}}{2019a}]%
        {hendrycks2019using}
\bibfield{author}{\bibinfo{person}{Dan Hendrycks}, \bibinfo{person}{Mantas
  Mazeika}, \bibinfo{person}{Saurav Kadavath}, {and} \bibinfo{person}{Dawn
  Song}.} \bibinfo{year}{2019}\natexlab{a}.
\newblock \showarticletitle{Using self-supervised learning can improve model
  robustness and uncertainty}. In \bibinfo{booktitle}{\emph{Advances in Neural
  Information Processing Systems}}. \bibinfo{pages}{15663--15674}.
\newblock


\bibitem[\protect\citeauthoryear{Hendrycks, Mu, Cubuk, Zoph, Gilmer, and
  Lakshminarayanan}{Hendrycks et~al\mbox{.}}{2020b}]%
        {hendrycks2019augmix}
\bibfield{author}{\bibinfo{person}{Dan Hendrycks}, \bibinfo{person}{Norman Mu},
  \bibinfo{person}{Ekin~D Cubuk}, \bibinfo{person}{Barret Zoph},
  \bibinfo{person}{Justin Gilmer}, {and} \bibinfo{person}{Balaji
  Lakshminarayanan}.} \bibinfo{year}{2020}\natexlab{b}.
\newblock \showarticletitle{Augmix: A simple data processing method to improve
  robustness and uncertainty}. In \bibinfo{booktitle}{\emph{Augmix: A simple
  data processing method to improve robustness and uncertainty (ICLR)}}.
\newblock


\bibitem[\protect\citeauthoryear{Hendrycks, Zhao, Basart, Steinhardt, and
  Song}{Hendrycks et~al\mbox{.}}{2019b}]%
        {hendrycks2019nae}
\bibfield{author}{\bibinfo{person}{Dan Hendrycks}, \bibinfo{person}{Kevin
  Zhao}, \bibinfo{person}{Steven Basart}, \bibinfo{person}{Jacob Steinhardt},
  {and} \bibinfo{person}{Dawn Song}.} \bibinfo{year}{2019}\natexlab{b}.
\newblock \showarticletitle{Natural Adversarial Examples}.
\newblock \bibinfo{journal}{\emph{arXiv preprint arXiv:1907.07174}}
  (\bibinfo{year}{2019}).
\newblock


\bibitem[\protect\citeauthoryear{Hosseini, Xiao, and Poovendran}{Hosseini
  et~al\mbox{.}}{2017}]%
        {hosseini2017google}
\bibfield{author}{\bibinfo{person}{Hossein Hosseini}, \bibinfo{person}{Baicen
  Xiao}, {and} \bibinfo{person}{Radha Poovendran}.}
  \bibinfo{year}{2017}\natexlab{}.
\newblock \showarticletitle{Google's cloud vision api is not robust to noise}.
  In \bibinfo{booktitle}{\emph{2017 16th IEEE International Conference on
  Machine Learning and Applications (ICMLA)}}. IEEE, \bibinfo{pages}{101--105}.
\newblock


\bibitem[\protect\citeauthoryear{Huang, Wang, Xing, and Huang}{Huang
  et~al\mbox{.}}{2020}]%
        {huang2020self}
\bibfield{author}{\bibinfo{person}{Zeyi Huang}, \bibinfo{person}{Haohan Wang},
  \bibinfo{person}{Eric~P. Xing}, {and} \bibinfo{person}{Dong Huang}.}
  \bibinfo{year}{2020}\natexlab{}.
\newblock \showarticletitle{Self-Challenging Improves Cross-Domain
  Generalization}. In \bibinfo{booktitle}{\emph{ECCV}}.
\newblock


\bibitem[\protect\citeauthoryear{Kang, Sun, Hendrycks, Brown, and
  Steinhardt}{Kang et~al\mbox{.}}{2019}]%
        {kang2019testing}
\bibfield{author}{\bibinfo{person}{Daniel Kang}, \bibinfo{person}{Yi Sun},
  \bibinfo{person}{Dan Hendrycks}, \bibinfo{person}{Tom Brown}, {and}
  \bibinfo{person}{Jacob Steinhardt}.} \bibinfo{year}{2019}\natexlab{}.
\newblock \showarticletitle{Testing robustness against unforeseen adversaries}.
\newblock \bibinfo{journal}{\emph{arXiv preprint arXiv:1908.08016}}
  (\bibinfo{year}{2019}).
\newblock


\bibitem[\protect\citeauthoryear{Li, Jiao, Cao, Wong, and Wu}{Li
  et~al\mbox{.}}{2020a}]%
        {li2020model}
\bibfield{author}{\bibinfo{person}{Rui Li}, \bibinfo{person}{Qianfen Jiao},
  \bibinfo{person}{Wenming Cao}, \bibinfo{person}{Hau-San Wong}, {and}
  \bibinfo{person}{Si Wu}.} \bibinfo{year}{2020}\natexlab{a}.
\newblock \showarticletitle{Model Adaptation: Unsupervised Domain Adaptation
  without Source Data}. In \bibinfo{booktitle}{\emph{Proceedings of the
  IEEE/CVF Conference on Computer Vision and Pattern Recognition}}.
  \bibinfo{pages}{9641--9650}.
\newblock


\bibitem[\protect\citeauthoryear{Li, Liu, Lin, Xie, Ding, Huang, and Tang}{Li
  et~al\mbox{.}}{2020b}]%
        {Li20DCAN}
\bibfield{author}{\bibinfo{person}{Shuang Li}, \bibinfo{person}{Chi~Harold
  Liu}, \bibinfo{person}{Qiuxia Lin}, \bibinfo{person}{Binhui Xie},
  \bibinfo{person}{Zhengming Ding}, \bibinfo{person}{Gao Huang}, {and}
  \bibinfo{person}{Jian Tang}.} \bibinfo{year}{2020}\natexlab{b}.
\newblock \showarticletitle{Domain Conditioned Adaptation Network}. In
  \bibinfo{booktitle}{\emph{Thirty-Fourth AAAI Conference on Artificial
  Intelligence (AAAI-20)}}.
\newblock


\bibitem[\protect\citeauthoryear{Liang, Hu, and Feng}{Liang
  et~al\mbox{.}}{2020}]%
        {liang2020shot}
\bibfield{author}{\bibinfo{person}{Jian Liang}, \bibinfo{person}{Dapeng Hu},
  {and} \bibinfo{person}{Jiashi Feng}.} \bibinfo{year}{2020}\natexlab{}.
\newblock \showarticletitle{Do We Really Need to Access the Source Data? Source
  Hypothesis Transfer for Unsupervised Domain Adaptation}. In
  \bibinfo{booktitle}{\emph{International Conference on Machine Learning
  (ICML)}}. \bibinfo{pages}{xx--xx}.
\newblock


\bibitem[\protect\citeauthoryear{Long, Cao, Wang, and Jordan}{Long
  et~al\mbox{.}}{2015}]%
        {long2015DAN}
\bibfield{author}{\bibinfo{person}{Mingsheng Long}, \bibinfo{person}{Yue Cao},
  \bibinfo{person}{Jianmin Wang}, {and} \bibinfo{person}{Michael Jordan}.}
  \bibinfo{year}{2015}\natexlab{}.
\newblock \showarticletitle{Learning transferable features with deep adaptation
  networks}. In \bibinfo{booktitle}{\emph{International conference on machine
  learning}}. PMLR, \bibinfo{pages}{97--105}.
\newblock


\bibitem[\protect\citeauthoryear{Long, Cao, Wang, and Jordan}{Long
  et~al\mbox{.}}{2018}]%
        {long2018conditional}
\bibfield{author}{\bibinfo{person}{Mingsheng Long}, \bibinfo{person}{Zhangjie
  Cao}, \bibinfo{person}{Jianmin Wang}, {and} \bibinfo{person}{Michael~I
  Jordan}.} \bibinfo{year}{2018}\natexlab{}.
\newblock \showarticletitle{Conditional adversarial domain adaptation}. In
  \bibinfo{booktitle}{\emph{Advances in Neural Information Processing
  Systems}}. \bibinfo{pages}{1640--1650}.
\newblock


\bibitem[\protect\citeauthoryear{Lopes, Yin, Poole, Gilmer, and Cubuk}{Lopes
  et~al\mbox{.}}{2019}]%
        {lopes2019improving}
\bibfield{author}{\bibinfo{person}{Raphael~Gontijo Lopes},
  \bibinfo{person}{Dong Yin}, \bibinfo{person}{Ben Poole},
  \bibinfo{person}{Justin Gilmer}, {and} \bibinfo{person}{Ekin~D Cubuk}.}
  \bibinfo{year}{2019}\natexlab{}.
\newblock \showarticletitle{Improving robustness without sacrificing accuracy
  with patch gaussian augmentation}.
\newblock \bibinfo{journal}{\emph{arXiv preprint arXiv:1906.02611}}
  (\bibinfo{year}{2019}).
\newblock


\bibitem[\protect\citeauthoryear{Madry, Makelov, Schmidt, Tsipras, and
  Vladu}{Madry et~al\mbox{.}}{2018}]%
        {madry2017PGD}
\bibfield{author}{\bibinfo{person}{Aleksander Madry},
  \bibinfo{person}{Aleksandar Makelov}, \bibinfo{person}{Ludwig Schmidt},
  \bibinfo{person}{Dimitris Tsipras}, {and} \bibinfo{person}{Adrian Vladu}.}
  \bibinfo{year}{2018}\natexlab{}.
\newblock \showarticletitle{Towards deep learning models resistant to
  adversarial attacks}. In \bibinfo{booktitle}{\emph{ICLR}}.
\newblock


\bibitem[\protect\citeauthoryear{Matsuura and Harada}{Matsuura and
  Harada}{2020}]%
        {dg_mmld}
\bibfield{author}{\bibinfo{person}{Toshihiko Matsuura} {and}
  \bibinfo{person}{Tatsuya Harada}.} \bibinfo{year}{2020}\natexlab{}.
\newblock \showarticletitle{Domain Generalization Using a Mixture of Multiple
  Latent Domains}. In \bibinfo{booktitle}{\emph{AAAI}}.
\newblock


\bibitem[\protect\citeauthoryear{Pan, Tsang, Kwok, and Yang}{Pan
  et~al\mbox{.}}{2010}]%
        {pan2010TCA}
\bibfield{author}{\bibinfo{person}{Sinno~Jialin Pan}, \bibinfo{person}{Ivor~W
  Tsang}, \bibinfo{person}{James~T Kwok}, {and} \bibinfo{person}{Qiang Yang}.}
  \bibinfo{year}{2010}\natexlab{}.
\newblock \showarticletitle{Domain adaptation via transfer component analysis}.
\newblock \bibinfo{journal}{\emph{IEEE Transactions on Neural Networks}}
  \bibinfo{volume}{22}, \bibinfo{number}{2} (\bibinfo{year}{2010}),
  \bibinfo{pages}{199--210}.
\newblock


\bibitem[\protect\citeauthoryear{Pan and Yang}{Pan and Yang}{2009}]%
        {pan2009survey}
\bibfield{author}{\bibinfo{person}{Sinno~Jialin Pan} {and}
  \bibinfo{person}{Qiang Yang}.} \bibinfo{year}{2009}\natexlab{}.
\newblock \showarticletitle{A survey on transfer learning}.
\newblock \bibinfo{journal}{\emph{IEEE Transactions on knowledge and data
  engineering}} \bibinfo{volume}{22}, \bibinfo{number}{10}
  (\bibinfo{year}{2009}), \bibinfo{pages}{1345--1359}.
\newblock


\bibitem[\protect\citeauthoryear{Saenko, Kulis, Fritz, and Darrell}{Saenko
  et~al\mbox{.}}{2010}]%
        {saenko2010office31}
\bibfield{author}{\bibinfo{person}{Kate Saenko}, \bibinfo{person}{Brian Kulis},
  \bibinfo{person}{Mario Fritz}, {and} \bibinfo{person}{Trevor Darrell}.}
  \bibinfo{year}{2010}\natexlab{}.
\newblock \showarticletitle{Adapting visual category models to new domains}. In
  \bibinfo{booktitle}{\emph{ECCV}}.
\newblock


\bibitem[\protect\citeauthoryear{Saito, Ushiku, and Harada}{Saito
  et~al\mbox{.}}{2017}]%
        {saito2017asymmetric}
\bibfield{author}{\bibinfo{person}{Kuniaki Saito}, \bibinfo{person}{Yoshitaka
  Ushiku}, {and} \bibinfo{person}{Tatsuya Harada}.}
  \bibinfo{year}{2017}\natexlab{}.
\newblock \showarticletitle{Asymmetric tri-training for unsupervised domain
  adaptation}.
\newblock \bibinfo{journal}{\emph{arXiv preprint arXiv:1702.08400}}
  (\bibinfo{year}{2017}).
\newblock


\bibitem[\protect\citeauthoryear{Sheng, Li, Zheng, Liang, Dong, Huang, Ji, and
  Sun}{Sheng et~al\mbox{.}}{2021}]%
        {sheng2021evolving}
\bibfield{author}{\bibinfo{person}{Kekai Sheng}, \bibinfo{person}{Ke Li},
  \bibinfo{person}{Xiawu Zheng}, \bibinfo{person}{Jian Liang},
  \bibinfo{person}{Weiming Dong}, \bibinfo{person}{Feiyue Huang},
  \bibinfo{person}{Rongrong Ji}, {and} \bibinfo{person}{Xing Sun}.}
  \bibinfo{year}{2021}\natexlab{}.
\newblock \showarticletitle{On Evolving Attention Towards Domain Adaptation}.
\newblock \bibinfo{journal}{\emph{arXiv preprint arXiv:2103.13561}}
  (\bibinfo{year}{2021}).
\newblock


\bibitem[\protect\citeauthoryear{Shorten and Khoshgoftaar}{Shorten and
  Khoshgoftaar}{2019}]%
        {shorten2019survey}
\bibfield{author}{\bibinfo{person}{Connor Shorten} {and}
  \bibinfo{person}{Taghi~M Khoshgoftaar}.} \bibinfo{year}{2019}\natexlab{}.
\newblock \showarticletitle{A survey on image data augmentation for deep
  learning}.
\newblock \bibinfo{journal}{\emph{Journal of Big Data}} \bibinfo{volume}{6},
  \bibinfo{number}{1} (\bibinfo{year}{2019}), \bibinfo{pages}{60}.
\newblock


\bibitem[\protect\citeauthoryear{Shu, Bui, Narui, and Ermon}{Shu
  et~al\mbox{.}}{2018}]%
        {shu2018dirt}
\bibfield{author}{\bibinfo{person}{Rui Shu}, \bibinfo{person}{Hung~H Bui},
  \bibinfo{person}{Hirokazu Narui}, {and} \bibinfo{person}{Stefano Ermon}.}
  \bibinfo{year}{2018}\natexlab{}.
\newblock \showarticletitle{A dirt-t approach to unsupervised domain
  adaptation}. In \bibinfo{booktitle}{\emph{ICLR}}.
\newblock


\bibitem[\protect\citeauthoryear{Sun, Wang, Zhuang, Miller, Hardt, and
  Efros}{Sun et~al\mbox{.}}{2020}]%
        {sun19ttt}
\bibfield{author}{\bibinfo{person}{Yu Sun}, \bibinfo{person}{Xiaolong Wang},
  \bibinfo{person}{Liu Zhuang}, \bibinfo{person}{John Miller},
  \bibinfo{person}{Moritz Hardt}, {and} \bibinfo{person}{Alexei~A. Efros}.}
  \bibinfo{year}{2020}\natexlab{}.
\newblock \showarticletitle{Test-Time Training with Self-Supervision for
  Generalization under Distribution Shifts}. In
  \bibinfo{booktitle}{\emph{ICML}}.
\newblock


\bibitem[\protect\citeauthoryear{Tsai, Hung, Schulter, Sohn, Yang, and
  Chandraker}{Tsai et~al\mbox{.}}{2018}]%
        {tsai2018learning}
\bibfield{author}{\bibinfo{person}{Yi-Hsuan Tsai}, \bibinfo{person}{Wei-Chih
  Hung}, \bibinfo{person}{Samuel Schulter}, \bibinfo{person}{Kihyuk Sohn},
  \bibinfo{person}{Ming-Hsuan Yang}, {and} \bibinfo{person}{Manmohan
  Chandraker}.} \bibinfo{year}{2018}\natexlab{}.
\newblock \showarticletitle{Learning to adapt structured output space for
  semantic segmentation}. In \bibinfo{booktitle}{\emph{Proceedings of the IEEE
  Conference on Computer Vision and Pattern Recognition}}.
  \bibinfo{pages}{7472--7481}.
\newblock


\bibitem[\protect\citeauthoryear{Vasiljevic, Chakrabarti, and
  Shakhnarovich}{Vasiljevic et~al\mbox{.}}{2016}]%
        {vasiljevic2016examining}
\bibfield{author}{\bibinfo{person}{Igor Vasiljevic}, \bibinfo{person}{Ayan
  Chakrabarti}, {and} \bibinfo{person}{Gregory Shakhnarovich}.}
  \bibinfo{year}{2016}\natexlab{}.
\newblock \showarticletitle{Examining the impact of blur on recognition by
  convolutional networks}.
\newblock \bibinfo{journal}{\emph{arXiv preprint arXiv:1611.05760}}
  (\bibinfo{year}{2016}).
\newblock


\bibitem[\protect\citeauthoryear{Venkateswara, Eusebio, Chakraborty, and
  Panchanathan}{Venkateswara et~al\mbox{.}}{2017}]%
        {venkateswara2017officehome}
\bibfield{author}{\bibinfo{person}{Hemanth Venkateswara}, \bibinfo{person}{Jose
  Eusebio}, \bibinfo{person}{Shayok Chakraborty}, {and}
  \bibinfo{person}{Sethuraman Panchanathan}.} \bibinfo{year}{2017}\natexlab{}.
\newblock \showarticletitle{Deep hashing network for unsupervised domain
  adaptation}. In \bibinfo{booktitle}{\emph{CVPR}}.
\newblock


\bibitem[\protect\citeauthoryear{Volpi, Namkoong, Sener, Duchi, Murino, and
  Savarese}{Volpi et~al\mbox{.}}{2018}]%
        {volpi2018generalizing}
\bibfield{author}{\bibinfo{person}{Riccardo Volpi}, \bibinfo{person}{Hongseok
  Namkoong}, \bibinfo{person}{Ozan Sener}, \bibinfo{person}{John~C Duchi},
  \bibinfo{person}{Vittorio Murino}, {and} \bibinfo{person}{Silvio Savarese}.}
  \bibinfo{year}{2018}\natexlab{}.
\newblock \showarticletitle{Generalizing to unseen domains via adversarial data
  augmentation}. In \bibinfo{booktitle}{\emph{Advances in neural information
  processing systems}}. \bibinfo{pages}{5334--5344}.
\newblock


\bibitem[\protect\citeauthoryear{Wang, Jin, Long, Wang, and Jordan}{Wang
  et~al\mbox{.}}{2019}]%
        {wang2019TransNorm}
\bibfield{author}{\bibinfo{person}{Ximei Wang}, \bibinfo{person}{Ying Jin},
  \bibinfo{person}{Mingsheng Long}, \bibinfo{person}{Jianmin Wang}, {and}
  \bibinfo{person}{Michael~I Jordan}.} \bibinfo{year}{2019}\natexlab{}.
\newblock \showarticletitle{Transferable normalization: Towards improving
  transferability of deep neural networks}. In
  \bibinfo{booktitle}{\emph{Advances in Neural Information Processing
  Systems}}. \bibinfo{pages}{1953--1963}.
\newblock


\bibitem[\protect\citeauthoryear{Yun, Han, Oh, Chun, Choe, and Yoo}{Yun
  et~al\mbox{.}}{2019}]%
        {yun2019cutmix}
\bibfield{author}{\bibinfo{person}{Sangdoo Yun}, \bibinfo{person}{Dongyoon
  Han}, \bibinfo{person}{Seong~Joon Oh}, \bibinfo{person}{Sanghyuk Chun},
  \bibinfo{person}{Junsuk Choe}, {and} \bibinfo{person}{Youngjoon Yoo}.}
  \bibinfo{year}{2019}\natexlab{}.
\newblock \showarticletitle{Cutmix: Regularization strategy to train strong
  classifiers with localizable features}. In
  \bibinfo{booktitle}{\emph{Proceedings of the IEEE International Conference on
  Computer Vision}}. \bibinfo{pages}{6023--6032}.
\newblock


\bibitem[\protect\citeauthoryear{Zhang, Cisse, Dauphin, and Lopez-Paz}{Zhang
  et~al\mbox{.}}{2018a}]%
        {zhang2017mixup}
\bibfield{author}{\bibinfo{person}{Hongyi Zhang}, \bibinfo{person}{Moustapha
  Cisse}, \bibinfo{person}{Yann~N Dauphin}, {and} \bibinfo{person}{David
  Lopez-Paz}.} \bibinfo{year}{2018}\natexlab{a}.
\newblock \showarticletitle{mixup: Beyond empirical risk minimization}.
\newblock \bibinfo{journal}{\emph{International Conference on Learning
  Representations}} (\bibinfo{year}{2018}).
\newblock


\bibitem[\protect\citeauthoryear{Zhang, Ouyang, Li, and Xu}{Zhang
  et~al\mbox{.}}{2018b}]%
        {zhang2018collaborative}
\bibfield{author}{\bibinfo{person}{Weichen Zhang}, \bibinfo{person}{Wanli
  Ouyang}, \bibinfo{person}{Wen Li}, {and} \bibinfo{person}{Dong Xu}.}
  \bibinfo{year}{2018}\natexlab{b}.
\newblock \showarticletitle{Collaborative and adversarial network for
  unsupervised domain adaptation}. In \bibinfo{booktitle}{\emph{Proceedings of
  the IEEE Conference on Computer Vision and Pattern Recognition}}.
  \bibinfo{pages}{3801--3809}.
\newblock


\bibitem[\protect\citeauthoryear{Zhang, Tang, Jia, and Tan}{Zhang
  et~al\mbox{.}}{2019}]%
        {zhang2019domain}
\bibfield{author}{\bibinfo{person}{Yabin Zhang}, \bibinfo{person}{Hui Tang},
  \bibinfo{person}{Kui Jia}, {and} \bibinfo{person}{Mingkui Tan}.}
  \bibinfo{year}{2019}\natexlab{}.
\newblock \showarticletitle{Domain-symmetric networks for adversarial domain
  adaptation}. In \bibinfo{booktitle}{\emph{Proceedings of the IEEE Conference
  on Computer Vision and Pattern Recognition}}. \bibinfo{pages}{5031--5040}.
\newblock


\end{thebibliography}

\end{document}

\title{Supplementary Material for ``Towards Corruption-Agnostic Robust Domain Adaptation''}

\author{Yifan Xu}
\affiliation{%
  \institution{NLPR, Institute of Automation, Chinese Academy of Sciences {\&} School of Artificial Intelligence, University of Chinese Academy of Sciences}
  \streetaddress{95 East Zhongguancun Rd}
  \city{Beijing}
  \country{China}
  \postcode{100190}
}
\email{yifan.xu@nlpr.ia.ac.cn}

\author{Kekai Sheng}
\affiliation{%
  \institution{Youtu Lab, Tencent Inc.}
  \city{Shanghai}
  \country{China}}
\email{saulsheng@tencent.com}

\author{Weiming Dong}
\affiliation{%
  \institution{ NLPR, Institute of Automation, Chinese Academy of Sciences {\&} CASIA-LLvision Joint Lab}
  \city{Beijing}
  \country{China}
}
\email{weiming.dong@ia.ac.cn}

\author{Baoyuan Wu}
\affiliation{%
 \institution{The Chinese University of Hong Kong, Shenzhen; Shenzhen Research Institute of Big Data}
 \city{ShenZhen}
 \country{China}}
\email{wubaoyuan1987@gmail.com}

\author{Changsheng Xu}
\affiliation{%
  \institution{NLPR, Institute of Automation, Chinese Academy of Sciences {\&} School of Artificial Intelligence, University of Chinese Academy of Sciences}
  \city{Beijing}
  \country{China}}
\email{csxu@nlpr.ia.ac.cn}

\author{Bao-Gang Hu}
\affiliation{%
  \institution{NLPR, Institute of Automation, Chinese Academy of Sciences}
  \city{Beijing}
  \country{China}}
\email{ hubg@nlpr.ia.ac.cn}

\renewcommand{\shortauthors}{Xu, et al.}


\maketitle
%\newpage
In this supplementary material, we provide details omitted in the main manuscript, including:
\begin{itemize}
    \item Section~\ref{sec:further_explanation}: Further explanation for the importance and challenge of the new task setting, CRDA.
    \item Section~\ref{sec:sup_ProofFor3Assumptions}: Empirical proof for three assumptions in Section 4.1 of the main manuscript.
    \item Section~\ref{sec:sup_ProofForEq6}: Proof for Equation (3) in the main manuscript.
    \item Section~\ref{sec:more_ablation}: More ablation study results.
    % \item Section~\ref{sec:full_TSCL_result}: Full results for the trade-off in TSCL in Section 5.2.
    \item Section~\ref{sec:sup_DetailsOfCorruptions}: Details of corruptions.
    % \item Section~\ref{sec:wc_sample}: Instance of worst-corruption samples.
    \item Section~\ref{sec:other_details}: More details of experiments.
\end{itemize}
%%%%%%%%% BODY TEXT
% %%%%%%%%% BODY TEXT - ENTER YOUR RESPONSE BELOW
\section{Further Explanation}
\label{sec:further_explanation}
% corruption robustness一直都很重要
% 以往研究不care这件事情是因为它在有监督的情境下已经解决的很好了
% 在DA中的corruption远远没有达到upper bound，还有非常大的工作空间
\begin{figure}[!h]
    \centering
    \includegraphics[width=0.6\linewidth]{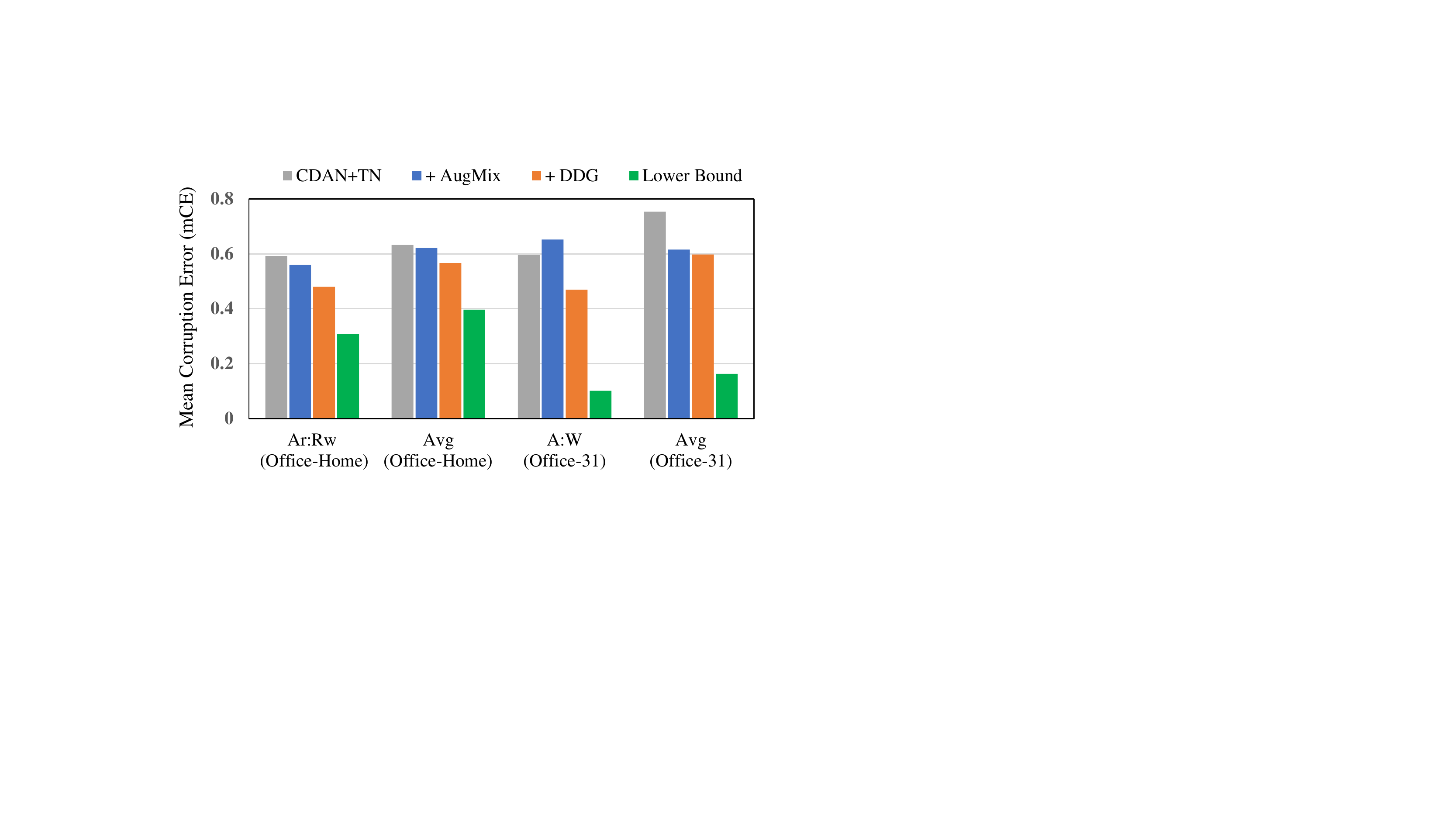}
    \caption{mCE ($\%$) on Office-Home and Office-31 datasets.}
    \label{fig:mCE-precise-sup}
\end{figure}

In this section, we emphasize the challenge and importance of CRDA through the lower bound. 

Corruption robustness is always an important subject. Previous works do not pay much care for corruptions because it has been addressed enough well in the setting of supervised learning. However, \textbf{corruption robustness in domain adaptation is far from satisfactory}.
% Corruption robustness has been tackled enough well in the setting of supervised learning, which is the reason why more attention are paid on challenging adversarial robustness rather than corruption robustness. However, corruption robustness in domain adaptation is far from satisfactory. 

As shown in Fig.~\ref{fig:mCE-precise-sup}, to estimate the status of current study, we further set an empirical lower bound for CRDA calculated by mCE of TSCL applied on CDAN+TN, which assumes testing corruptions are available while training. Models that reach the lower bound can be robust against unseen corruptions as if they are already known beforehand.

It is shown that current methods for CRDA still hold a large distance to the lower bound. In a word, instead of enough good progress in supervised learning, \textbf{there is still a long way to go for CRDA.}

%----------------------------------------------------------------
\section{Proof for three assumptions in Section 4.1 of the main manuscript}
\label{sec:sup_ProofFor3Assumptions}

% \begin{assumption}
% \label{assum:assum1}
% The corrupted versions are within the $\delta$-range neighborhood of the original clean image in sample space ($\boldsymbol{\delta}$ \textbf{cycle} for precise) and distance in sample space $d_{s}(t(x),x)$ becomes larger with increase of the severity level of corruption $T$.
% \end{assumption}
% \begin{assumption}
% \label{assum:assum2}
% Within the $\delta$ cycle, the severity level $t$ of a corruption  $T$ is positively correlated to distance in feature space  $d_{f}(f(t(x)) ,f(x))$.
% \end{assumption}
% \begin{assumption}
% \label{assum:assum3}
% Within the $\delta$ cycle, the severity level $t$ of a corruption $T$ is positively correlated to the value of transfer loss $\ell_{trans}(t(x))$.
% \end{assumption}

%--------------------------------------------------------------------------
\subsection{Proof for Assumption 1}
We give the empirical proof for Assumption 1 via Average Shift. Given a clean image $x\in \mathbb{R}^{c \times w \times h}$ and a corruption $t$, the Average Shift is calculated by:
\begin{equation}
    \begin{split}
    avg&=\frac{\|x-t(x)\|_{1}}{cwh} \\
    &=\frac{1}{cwh}\sum_{i=1}^{c} \sum_{j=1}^{w} \sum_{k=1}^{h} \left|(x-t(x))_{i j k}\right|.
    \end{split}
\end{equation}

Table \ref{tab:shift} shows the Average Shift of an image under 15 corruptions with the most severe level. It is shown that most corruptions are within the shift range $\delta=0.26 \approx 60/255$.

\begin{figure}[t]
    \centering
    \includegraphics[width=0.45\textwidth]{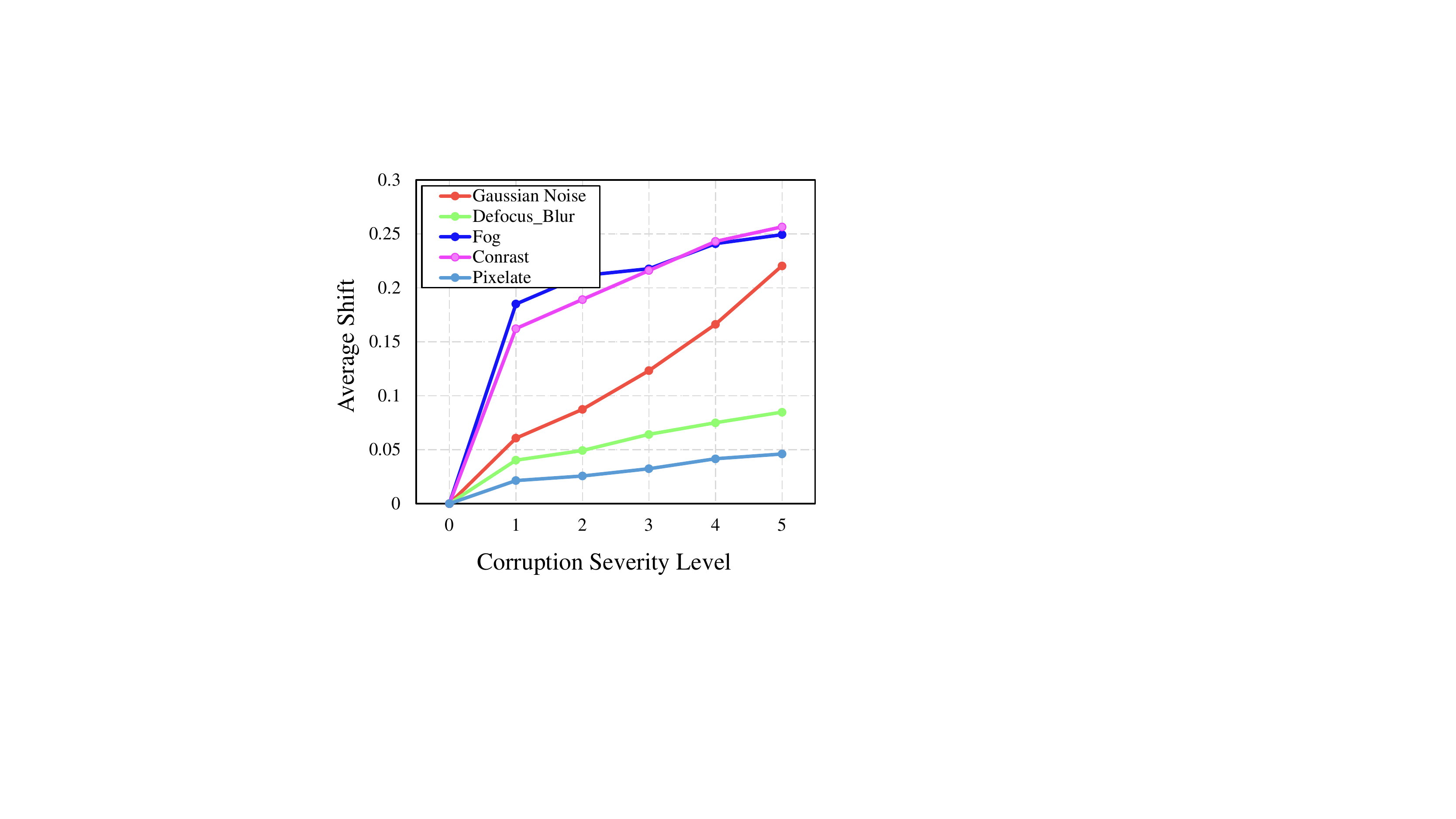}
    \caption{Average Shift of different corruptions with $6$ severity levels.}
    \label{fig:shift-6severity}
\end{figure}

Fig.~\ref{fig:shift-6severity} verifies that the Average Shift in sample space increases with the increase of the severity level of most corruptions.

\begin{table}[t]
    \centering
    \caption{Average Shift of different corruptions}
    \label{tab:shift}
    \setlength{\tabcolsep}{1.2mm}{
    \small{
    \begin{tabular}{c|c||c|c}
        \toprule
         Corruption & Avg shift & Corruption & Avg shift \\
         \hline
         Gaussian Noise & 0.22 & Snow & 0.25\\
         Shot Noise & 0.25 & Elastic Transform & 0.19\\
         Impulse Noise & 0.14 & Contrast & 0.26\\
         Defocus Blur & 0.08 & Brightness & 0.25\\
         Motion Blur & 0.13 & JPEG Compression & 0.04\\
         Zoom Blur & 0.12 & Pixelate & 0.05\\
         Fog & 0.24 & Glass Blur & 0.08\\
         Frost & 0.21 & & \\
         \bottomrule
    \end{tabular}
    }
    }
\end{table}

% \begin{table}
%     \centering
%     \caption{mCE ($\%$) on all cross-domain settings of Office-Home dataset under CRDA (ResNet-50).}
%     \label{tab:mCE_OfficeHome}
%     \setlength{\tabcolsep}{1.2mm}{
%     \small{
%     \begin{tabular}{c|ccc|ccc|c}
%         \toprule
%         \multirow{2}{*}{Settings} & \multicolumn{3}{c|}{CDAN+TN} &\multicolumn{3}{c|}{DCAN} & Lower \\
%         \cline{2-7}
%          & - & AugMix & Ours & - & AugMix & Ours & Bound \\
%         \hline
%         Ar$\to$Cl & 69.7 & 69.3 & 59.8 & 61.4 & 63.2 & 58.3 & 52.2 \\
%         Ar$\to$Pr & 62.2 & 52.8 & 52.2 & 51.0 & 49.3 & 48.0 & 37.7 \\
%         Ar$\to$Rw & 59.2 & 56.0 & 48.0 & 47.9 & 46.0 & 44.6 & 30.8 \\
%         Cl$\to$Ar & 65.4 & 73.7 & 66.0 & 58.6 & 57.4 & 55.2 & 45.8 \\
%         Cl$\to$Pr & 58.7 & 56.6 & 51.9 & 58.0 & 57.3 & 51.6 & 34.6 \\
%         Cl$\to$Rw & 57.6 & 61.3 & 56.2 & 52.0 & 53.8 & 47.5 & 34.7 \\
%         Pr$\to$Ar & 65.5 & 72.2 & 63.6 & 61.1 & 58.9 & 55.1 & 44.7 \\
%         Pr$\to$Cl & 70.9 & 71.4 & 63.6 & 66.2 & 65.3 & 62.2 & 54.9 \\
%         Pr$\to$Rw & 55.0 & 55.3 & 48.8 & 53.7 & 48.5 & 49.4 & 25.7 \\
%         Rw$\to$Ar & 63.2 & 62.4 & 57.8 & 63.0 & 59.2 & 57.7 & 36.0 \\
%         Rw$\to$Cl & 70.4 & 61.4 & 64.4 & 64.0 & 64.2 & 59.9 & 52.9 \\
%         Rw$\to$Pr & 60.9 & 52.6 & 47.9 & 58.4 & 52.5 & 52.6 & 26.8 \\
%         \hline
%         Avg ($\downarrow$) & 63.2 & 62.1 & \textbf{56.7} & 57.9 & 56.3 & \textbf{53.5} & 39.7 \\
%         \bottomrule
%     \end{tabular}
%     }
%     }
%     \vspace{-1.5mm}
% \end{table}

%--------------------------------------------------------------------------
\subsection{Proof for Assumption 2}
It is shown in Section 5.4 of the main manuscript.

%------------------------------------------------------------------------
\subsection{Proof for Assumption 3}
Given input target domain $D_{t}=\{(\mathcal{X}_{t},\mathcal{Y}_{t})\}$, source domain $D_{s}=\{(\mathcal{X}_{s},\mathcal{Y}_{s})\}$, a corruption with 5 severity levels $T={\{t_{i}\}}_{1}^{5}$, an ideal transfer loss function $\ell_{trans}$ and a DA model $m=f \circ c$, we first make some notions for precise:

\begin{equation}
    \begin{split}
        \ell_{trans}(t_{i})&=\mathbb{E}_{x_t \sim \mathcal{X}_{t}} \left[\ell_{trans}(f(t_{i}(x_t)),f(\mathcal{X}_{s}))\right],\\
        Acc({t_{i}})&=\mathbb{P}_{(x_{t},y_{t}) \sim D_{t}}(m(t_{i}(x_{t}))=y_{t}),\\
        \ell_{total}(t_{i})&=\ell_{trans}(t_{i})+\ell_{cls}(\mathcal{X}_s,\mathcal{Y}_s), 
    \end{split}
\end{equation}
where $\ell_{trans}(t_{i})$ denotes the average transfer loss of corrupted target domain samples. $Acc({t_{i}})$ denotes the classification accuracy of corrupted target domain samples. $\ell_{cls}(\mathcal{X}_s)$ denotes the classifier loss of source domain samples; $\ell_{total}(t_{i})$ denotes the total loss of a classic DA models, which usually correlated to $Acc({t_{i}})$.

It is observed that the classification accuracy decreases as the increase of the severity level, as:
\begin{equation}
    i<j \Rightarrow Acc({t_{i}}) > Acc({t_{j}}).
    \label{observe}
\end{equation}

We make the proof by contradiction.
If
\begin{equation}
    \begin{split}
        i<j &\Rightarrow \ell_{trans}({t_{i}}) > \ell_{trans}({t_{j}})\\
        &\Rightarrow \ell_{total}(t_{i}) > \ell_{total}(t_{j}), 
    \end{split}
\end{equation}
the final classification accuracy should be $Acc({t_{i}}) < Acc({t_{j}})$. Conflict with Equation (\ref{observe}).

%----------------------------------------------------------------
\section{Proof for Equation (3) in the main manuscript}
\label{sec:sup_ProofForEq6}
% \begin{equation}
%     \begin{split}
%     x_{t}^{(DDG)}&=\mathop{\arg\max}_{\left\|x_{t}^{(DDG)}-x_{t}\right\| \leq \delta} \ell_{trans}\left(f(x_{t}^{(DDG)}), f(D_{s})\right)
%     \\&\Rightarrow \mathop{\arg\max}_{t \in T} \ell_{trans}\left(f(t\left(x_{t})\right), f(D_{s})\right)=t_{\max }(x_{t}).
%     \end{split}
%     \label{construction_x_adv}
% \end{equation}

According to Assumption 1, $x_{t}^{(DDG)}$ and $t(x_{t})$ are within the $\delta$ neighborhood $\{x | \left\|x-x_{t}\right\| \leq \delta\}$. 

Suppose $t_{max}$ as the most severe corruption in corruption set $T$. According to Assumption 3, 

\begin{equation}
   t_{max}=\mathop{\arg\max}_{t \in T} \ell_{trans}\left(f(t\left(x_{t})\right), f(D_{s})\right).
\end{equation}

Due to the uncertainty of $T$, the corrupted versions $t(x_{t})$ can be everywhere in $\delta$ neighborhood.
Thus, 
% $\mathop{\max}_{\left\|x_{t}^{(DDG)}-x_{t}\right\| \leq \delta} \ell_{trans}\left(f(x_{t}^{(DDG)}), f(D_{s})\right) = \mathop{\max}_{t \in T} \ell_{trans}\left(f(t\left(x_{t})\right), f(D_{s})\right)$
% 
% , as : 
\begin{equation}
    \begin{split}
    & \ell_{trans}\left(f(x_{t}^{(DDG)}), f(D_{s})\right)
    \\
    &=\ell_{trans}\left(f(t_{max}\left(x_{t})\right), f(D_{s})\right)=\ell_{max}.
    \end{split}
\end{equation}

Suppose that

\begin{equation}
    \begin{split}
     X^{(DDG)}&=\left \{x_{t}^{(DDG)}| \ell_{trans}\left(f(x_{t}^{(DDG)}), f(D_{s})\right)= \ell_{max} \right\}
    \\X^{(tm)}&=\left\{t_{max}(x_{t}) | \ell_{trans}\left(f(t_{max}(x_{t})), f(D_{s})\right) = \ell_{max} \right\}.
    \end{split}
\end{equation}
We get
\begin{equation}
    X^{(DDG)}=X^{(tm)}.
\end{equation}
Thus, $x_{t}^{(DDG)} \in X^{(wc)}$ can represent $t_{max}(x_{t}) \in X^{(tm)}$, as:

\begin{equation}
    x_{t}^{(DDG)} \Rightarrow t_{max}(x_{t}).
\end{equation}
%----------------------------------------------------------------
\section{More ablation on $\delta$, $\eta$, and $n$}
\label{sec:more_ablation}
% \begin{table}[t]
%     \centering
%     \caption{mCE ($\%$) on Ar$\to$Rw for different parameters in TSCL-WC}
%     \label{tab:Ablation-delta2}
%     \small{
%     \setlength{\tabcolsep}{4mm}{
%     \begin{tabular}{c|ccc|c|c}
%         \toprule
%         object&$\delta$&$\eta$&$n$&mCE ($\downarrow$)&convergence\\
%         \hline
%         \multirow{4}{*}{$\eta$} & $60/255$ & $> 2\delta$ & 2 & 48.0 & 2000\\ 
%         &$60/255$ & $> 2\delta$ & 10 & 49.6 & 3500\\ 
%         &$60/255$ & $15/255$ & 10 & 51.6 & 4000\\ 
%         &$60/255$ & $6/255$ & 10 & 52.8 & 4000\\ 
%         \hline
%         \multirow{7}{*}{$\delta$}&$20/255$ & $> 2\delta$ & 2 & 49.5 & 2000\\ 
%         &$40/255$ & $> 2\delta$ & 2 & 48.6 & 2000\\ 
%         &$60/255$ & $> 2\delta$ & 2 & 48.0 & 2000\\ 
%         &$80/255$ & $> 2\delta$ & 2 & 47.3 & 2500\\
%         &$100/255$ & $> 2\delta$ & 2 & 48.1 & 2500\\
%         &$120/255$ & $> 2\delta$ & 2 & 46.9 & 3000\\
%         &$140/255$ & $> 2\delta$ & 2 & 46.4 & 3000\\
%         \hline
%         \multirow{4}{*}{$n$}&$60/255$ & $> 2\delta$ & 1 & 47.4 & 2000\\
%         &$60/255$ & $> 2\delta$ & 2 & 48.0 & 2000\\
%         &$60/255$ & $> 2\delta$ & 5 & 49.9 & 2000\\
%         &$60/255$ & $> 2\delta$ & 10 & 49.6 & 3500\\
%         \bottomrule
%     \end{tabular}
%     }
%     }
%     \vspace{-3mm}
% \end{table}

\begin{table}[t]
    \centering
    \caption{mCE ($\%$) for different parameters in TSCL-WC on the UDA task of Ar$\to$Rw from Office-Home dataset.}
    \label{tab:Ablation-delta2}
    % \small{
    \setlength{\tabcolsep}{3mm}{
    \begin{tabular}{c|ccc|c}
        \toprule
        object&$\delta$&$\eta$&$n$&mCE ($\downarrow$)\\
        \hline
        \multirow{4}{*}{$\eta$} & $60/255$ & $> 2\delta$ & 2 & 48.0\\ 
        &$60/255$ & $> 2\delta$ & 10 & 49.6\\ 
        &$60/255$ & $15/255$ & 10 & 51.6\\ 
        &$60/255$ & $6/255$ & 10 & 52.8\\ 
        \hline
        \multirow{7}{*}{$\delta$}&$20/255$ & $> 2\delta$ & 2 & 49.5\\ 
        &$40/255$ & $> 2\delta$ & 2 & 48.6\\ 
        &$60/255$ & $> 2\delta$ & 2 & 48.0\\ 
        &$80/255$ & $> 2\delta$ & 2 & 47.3\\
        &$100/255$ & $> 2\delta$ & 2 & 47.2\\
        &$120/255$ & $> 2\delta$ & 2 & 46.9\\
        &$140/255$ & $> 2\delta$ & 2 & 46.4\\
        \hline
        \multirow{4}{*}{$n$}&$60/255$ & $> 2\delta$ & 1 & 47.4\\
        &$60/255$ & $> 2\delta$ & 2 & 48.0\\
        &$60/255$ & $> 2\delta$ & 5 & 49.9\\
        &$60/255$ & $> 2\delta$ & 10 & 49.6\\
        \bottomrule
    \end{tabular}
    % }
    }
    % \vspace{-3mm}
\end{table}
Table \ref{tab:Ablation-delta2} reports the mCE on different settings of hyper-parameter of DDG in Algorithm 2. Besides the conclusion of $\eta$ in the main text, it is shown that the final corruption robustness improves as the increase of shift range $\delta$ and decrease of the update step $n$. For stability, we set $\delta=60/255$, $\eta>2\delta$ and $n=2$ in this paper.
%---------------------------------------------------------------
% \input{latex/Supplementary_Section/Full_TSCL_results}
%--------------------------------------------------------------
\section{Details of corruptions}
\label{sec:sup_DetailsOfCorruptions}

\begin{figure*}[h]
    \centering
    \includegraphics[width=0.85\textwidth]{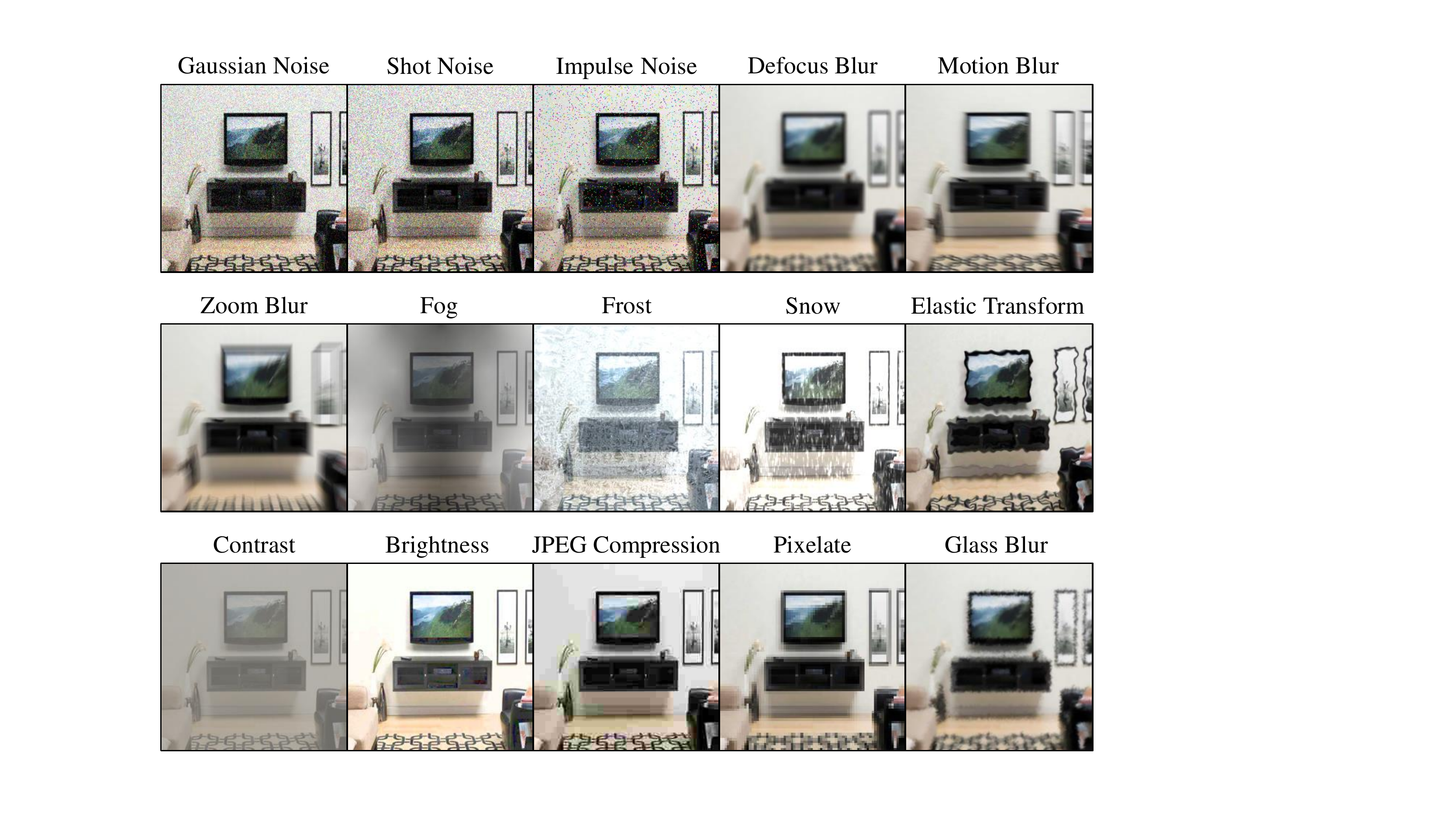}
    \caption{Examples of 15 kinds of corruptions.}
    \label{fig:15corruption}
\end{figure*}

\begin{figure*}[h]
    \centering
    \includegraphics[width=0.85\textwidth]{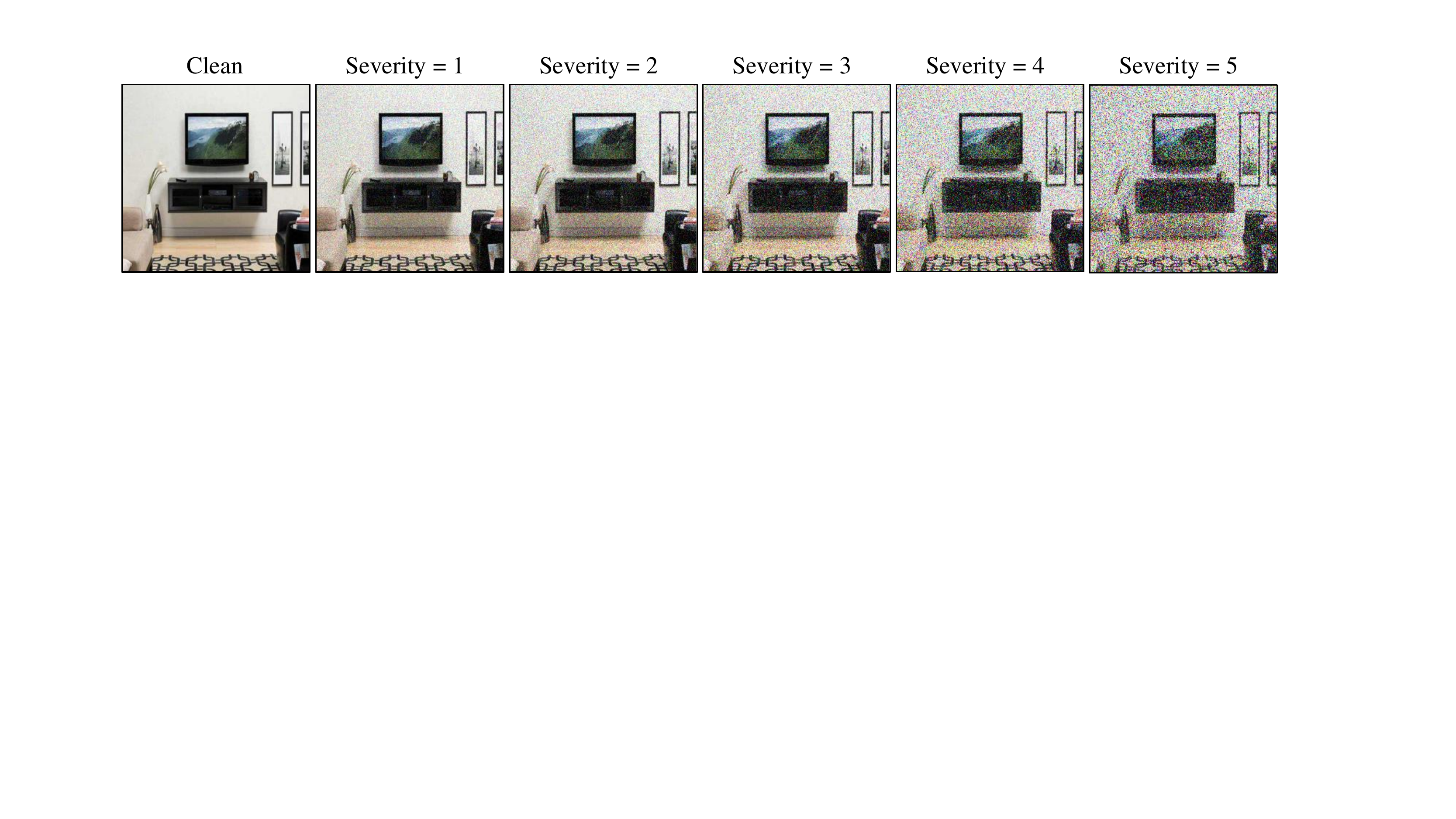}
    \caption{Examples of 5 levels of severity.}
    \label{fig:6severity}
\end{figure*}

We follow the same types of corruptions as Hendrycks~\etal~\cite{hendrycks2019benchmarking} proposed to investigate how neural networks are robust against to common corruptions and perturbations.
The $15$ types of corruptions are: Gaussian Noise, Shot Noise, Impluse Noise, Defocus Blur, Motion Blur, Zoom Blur, Fog, Frost, Snow, Elastic Transform, Contrast, Brightness, JPEG Compression, Pixelate, and Glass Blur.
Fig.~\ref{fig:15corruption} illustrates examples of all the $15$ kinds of corruptions. For each kind, there are $5$ levels of severity as shown in Fig.~\ref{fig:6severity}.

% \input{latex/Supplementary_Section/imgs/15corruptions}

% \input{latex/Supplementary_Section/imgs/6severity}
%----------------------------------------------------------------
% \input{latex/Supplementary_Section/Worst-Corruption_Sample}
%----------------------------------------------------------------
\section{Other details}
\label{sec:other_details}
% \paragraph{Clean error rate in Table 3 of Section 5.2.}
% The error rate of clean data in Table 3 is computed by the average of TSCL on total 15 corruptions (TSCL is trained for a specific corruption).

\paragraph{Training details of Fig.5 of Section 5.3 in the main manuscript}
All models are trained on a single domain (Real World of Office-Home).
Given Gaussian Noise corruption set with five levels of severity $T=\{t_{i}\}_{1}^{5}$ ($t_{0}$ denotes clean images) and inputs $x$, the training details of the models are as follows.
\begin{itemize}
    \item \textbf{Clean:} Supervised training on clean data.
    \item \textbf{All levels:} At the begining of each iteration, we randomly select a corruption from $\{t_{i}\}_{0}^{5}$ to corrupt the inputs $x$. All corrupted inputs $t(x)$ are aligned with the original clean inputs $x$ in feature space by contrastive loss.
    \item \textbf{Clean + Level 5:}  At the begining of each iteration, we randomly select a corruption from $\{t_{0},t_{5}\}$ to corrupt the inputs $x$. All corrupted inputs $t(x)$ are aligned with the original clean inputs $x$ in feature space by contrastive loss.
    \item \textbf{Level 5:} At the beginning of each iteration, we use level 5 corruption $t_{5}$ to corrupt the inputs $x$.
\end{itemize}

%----------------------------------------------------------------
% \input{latex/Supplementary_Section/imgs/wc_sample}

\bibliographystyle{ACM-Reference-Format}
\bibliography{references}